\documentclass[conference]{IEEEtran}
\IEEEoverridecommandlockouts
\usepackage{cite}
\usepackage{amsmath,amssymb,amsfonts}
\usepackage{algorithmic}
\usepackage{graphicx}
\usepackage{textcomp}
\usepackage{xcolor}
\usepackage{enumitem} 
\usepackage{stfloats}
\usepackage{placeins}
\usepackage{todonotes}
\usepackage{hyperref}
\usepackage{tabularx}
\usepackage{multicol}
\usepackage{array,multirow}
\usepackage{pgfplots}
\usepackage{pgfplotstable}
\usepackage{xcolor}
\usepackage{subcaption}
\usetikzlibrary{patterns}
\usepackage{booktabs}

\def\BibTeX{{\rm B\kern-.05em{\sc i\kern-.025em b}\kern-.08em
    T\kern-.1667em\lower.7ex\hbox{E}\kern-.125emX}}
\begin{document}

\title{Fair Recourse for All: Ensuring Individual and Group Fairness in Counterfactual Explanations

\thanks{Fatima Ezzeddine and Obaida Ammar are supported by the Swiss Government Excellence Scholarships.}
}

\author{\IEEEauthorblockN{Fatima Ezzeddine\textsuperscript{†,*}, Obaida Ammar\textsuperscript{†,*}, Silvia Giordano\textsuperscript{†} and Omran Ayoub\textsuperscript{†}}
\IEEEauthorblockA{
\textit{\textsuperscript{†} University of Applied Sciences and Arts of Southern Switzerland, Lugano, Switzerland}\\
\textit{\textsuperscript{*} Università della Svizzera italiana, Lugano, Switzerland}
}
}

\maketitle

\begin{abstract}
Explainable Artificial Intelligence (XAI) is becoming increasingly essential for enhancing the transparency of machine learning (ML) models. Among the various XAI techniques, counterfactual explanations (CFs) hold a pivotal role due to their ability to illustrate how changes in input features can alter an ML model's decision, thereby offering actionable recourse to users. Ensuring that individuals with comparable attributes and those belonging to different protected groups (e.g., demographic) receive similar and actionable recourse options is essential for trustworthy and fair decision-making. In this work, we address this challenge directly by focusing on the generation of fair CFs. Specifically, we start by defining and formulating fairness at: 1) individual fairness, ensuring that similar individuals receive similar CFs, 2) group fairness, ensuring equitable CFs across different protected groups and 3) hybrid fairness, which accounts for both individual and broader group-level fairness. We formulate the problem as an optimization task and propose a novel model-agnostic, reinforcement learning based approach to generate CFs that satisfy fairness constraints at both the individual and group levels, two objectives that are usually treated as orthogonal. As fairness metrics, we extend existing metrics commonly used for auditing ML models, such as equal choice of recourse and equal effectiveness across individuals and groups. We evaluate our approach on three benchmark datasets, showing that it effectively ensures individual and group fairness while preserving the quality of the generated CFs in terms of proximity and plausibility, and quantify the cost of fairness in the different levels separately. Our work opens a broader discussion on hybrid fairness and its role and implications for XAI and beyond CFs.
\end{abstract}


\section{Introduction}
Machine learning (ML) continues to see widespread adoption across a wide range of fields, spanning high-stakes domains such as healthcare, finance, and transportation, enabling data-driven insights and facilitating decision-making processes and predictive modeling. For high-risk applications such as, e.g., criminal justice, there exists a concern relative to the transparency of the deployed ML models, which are frequently perceived as black boxes of opaque nature offering little to no insights into their internal workings. This opacity hinders the ability to understand how these systems arrive at their decisions \cite{miller2019explanation}. Enhancing transparency is essential, on one hand, it allows practitioners to debug the model's reasoning and their logical process, with the final aim of pinpointing potential undesired biases and validating results. On the other hand, transparency allows end-users and stakeholders to trust the reasoning behind automated decisions \cite{gryz2021black}.

To this end, Explainable AI (XAI) techniques have recently gained attraction as a crucial area of research \cite{angelov2021explainable}. Specifically, XAI techniques provide explanations of a model behavior that come in different forms, such as feature attribution summaries and Counterfactual Explanations (CFs) \cite{wachter2017counterfactual}. CFs have gained specific interest due to them retaining the fidelity of the underlying ML model, thereby providing reliable insights into its decision logic \cite{wachter2017counterfactual} and providing recourse \cite{keane2021if}, which refers to an individual's ability to take actions to achieve a desired outcome. Specifically, CFs provide ``what-if" scenarios that illustrate which modifications to the input could lead to alternative outcomes and actionable insights into the decision-making process of ML models.

\begin{figure*}[htbp]
    \centering
    \begin{subfigure}[b]{0.32\textwidth}
        \includegraphics[width=\textwidth]{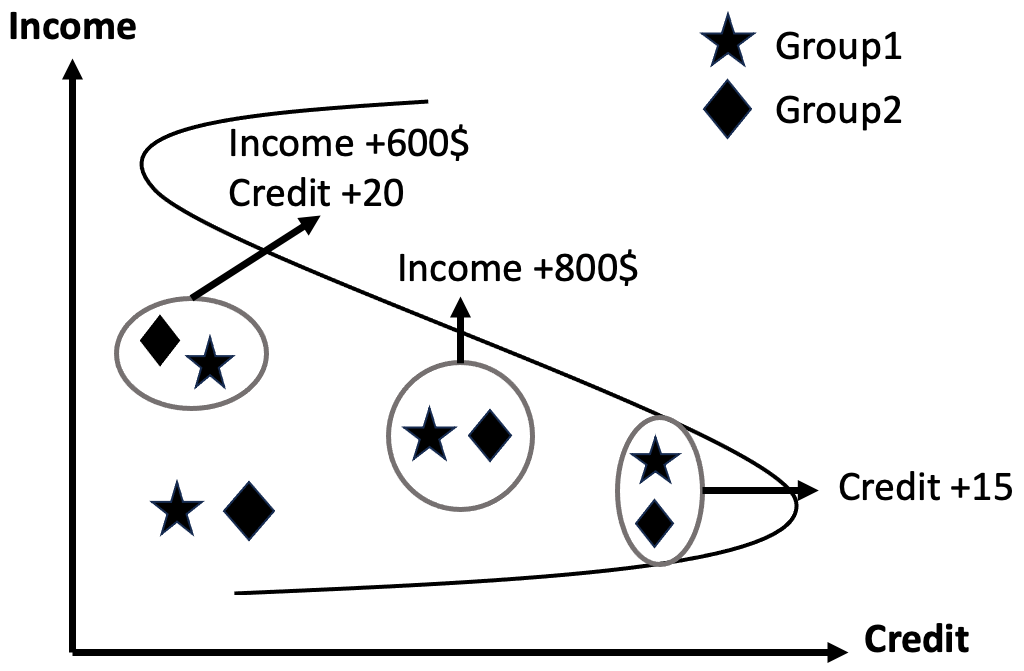}
        \caption{Equal Effectiveness-Individual}
        \label{fig:EF-Micro}
    \end{subfigure}
    \begin{subfigure}[b]{0.32\textwidth}
        \includegraphics[width=\textwidth]{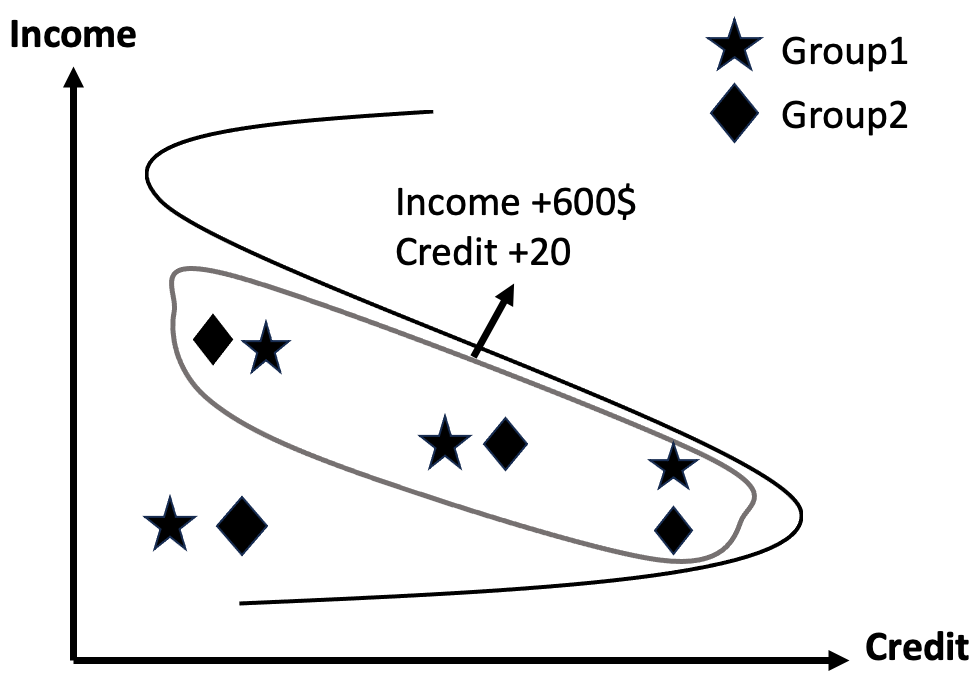}
        \caption{Equal Effectiveness-Group}
        \label{fig:EF-Macro}
    \end{subfigure}
    \begin{subfigure}[b]{0.32\textwidth}
        \includegraphics[width=\textwidth]{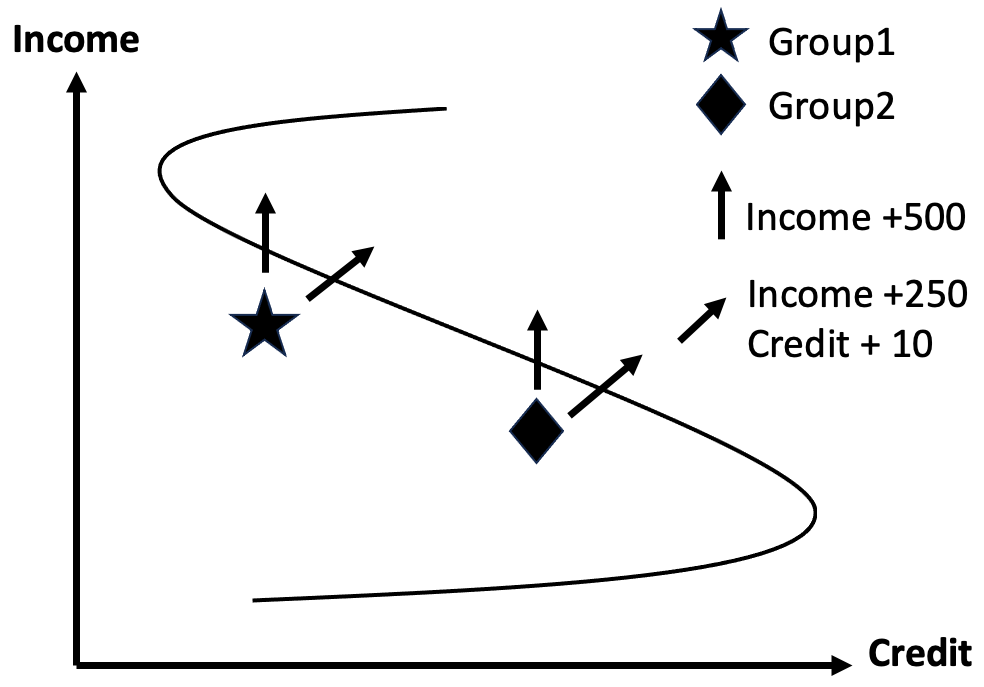}
        \caption{Equal choice of recourse-Group}
        \label{fig:ECR}
    \end{subfigure}
    \caption{Illustration of Equal Effectiveness (Individual and Group) and Equal Choice for Recourse}
    \label{fig:EF(Micro-Macro)-ECR}
\end{figure*}

Beyond technical transparency, fairness is indispensable to protect individuals from discriminatory treatment and to ensure equitable outcomes in ML–driven decisions. Indeed, fairness is a concern that must be addressed at both the model level and the explanation level \cite{artelt2022explain}. At the model level, fairness is violated when two individuals with comparable characteristics receive different outcomes due to attributes that should not influence decisions, such as protected attributes like gender or race \cite{pessach2022review,shaham2025privacy}. For instance, if two individuals have comparable financial profiles but different genders, for one, the model decides to approve a loan, while for the other, it does not. This may occur due to biased training data, which may reflect inequalities, such as individuals from certain demographic groups. Models trained on such data may internalize these associations, resulting in biased decisions for specific groups.

Ensuring fairness in explanations goes beyond the fairness of the underlying ML model, as the methods used to generate explanations can themselves differ and potentially propagate or amplify unfairness \cite{zhao2022fairnessexplainabilitybridginggap,Foulds2019DifferentialF,Bird2020FairlearnAT}. Even minor feature perturbations can compromise the fairness of CF recourse by limiting its consistency and similarity across comparable individuals from different groups \cite{artelt2021evaluating,artelt2022explain}, as authors in \cite{slack2021counterfactual} show that minor data perturbations can result in up to 20 times higher recourse costs for marginalized groups. For instance, consider two individuals with comparable financial profiles who are both denied a loan by an ML model. For the first individual, the CF suggests increasing the salary by 10k. For the second individual, the CF advises adding a co-signer and improving their credit score, without mentioning salary. This discrepancy, where individuals in similar circumstances receive different CFs, indicates the presence of bias. Such bias can be attributed to the fact that CF generation methods aim to find the smallest possible change needed to alter the prediction, however, if the smallest change for individuals within a particular group is to modify an immutable or disproportionately difficult feature, such as age or educational background, the resulting CFs become not only impractical but also unfair. Biased CFs thus have consequences beyond a single explanation, as they erode user confidence and undermine trust not only in the AI system but also in the organization deploying it, as the reliance on biased AI systems can result in reputational harm and legal liability.

In this context, ensuring fairness in CFs, a problem formally referred to as Fair CF, is equally critical to generating a traditional CF \cite{kuratomi2022measuring,Gupta2019EqualizingRA}. A fair CF generation method should explicitly integrate notions of fairness into its optimization objectives. Fairness in CFs (discussed formally in Sec. \ref{Preliminaries}) focuses on ensuring that similar individuals (i.e., have similar attributes) and individuals pertaining to different demographic groups have comparable explanations. 

Fair CFs can be considered at the individual level, group level, or, as we will show later, at a hybrid level, involving both individuals and the group they pertain to. More specifically, individual fairness \cite{li2023fairness,artelt2021evaluating} refers to treating individuals with comparable attributes similarly. Group fairness \cite{artelt2022explain,Bynum2024ANP} aims to ensure fairness at a broader level, focusing on the equal treatment of different groups to avoid discrimination based on group-sensitive attributes such as, e.g., gender or race. Individual and group fairness in ML often exhibit a complex, contradictory, and orthogonal relationship as each has a specific focus and aims \cite{li2023fairness}. Recently, the notion of \emph{hybrid fairness} emerged \cite{li2023fairness,deldjoo2024fairness}, where, instead of prioritizing a single fairness consideration (individual or group), it acknowledges the diverse and often conflicting nature of fairness demands, and aims to achieve multiple fairness requirements simultaneously. On the model level, the research in \cite{xu2024compatibility} explores the conditions under which group fairness and individual fairness can be achieved simultaneously. It provides theoretical analyses and methods to balance the trade-offs between these levels of fairness for recommender systems. On the explanation level, and more specifically, in the context of CF explanations, hybrid fairness indicates that the explanations provided are not only individually fair, giving comparable explanations to similar individuals with a comparable set of attributes, but also collectively fair, avoiding disparate impacts across different demographic groups. To date, little to no work has specifically explored approaches that aim to balance group-level and individual-level fairness in generating CFs.


In this paper, we address the problem of fair CF generation considering individual, group, and hybrid fairness by first defining metrics that capture fairness at these levels,  and then formulating the problem as a model-agnostic optimization task solved using reinforcement learning (RL). We then customize reward functions that account for fairness, such as optimizing for the proportion of individuals and groups achieving recourse, diversity of choices for groups, and quality of CFs in terms of common metrics such as plausibility, actionability, and proximity. Specifically, to take fairness into account, we shift the definitions of common properties such as equal effectiveness (EE) and equal choice of recourse (ECR), which were originally introduced to audit fairness in ML models \cite{kavouras2024fairness}, and use them as guiding principles for the RL. We focus on \emph{EE (Individual Fig. \ref{fig:EF-Micro}, Group Fig. \ref{fig:EF-Macro} and \ref{fig:ECR})} that indicate that individuals across different demographic groups receive comparable CFs when seeking to achieve a favorable outcome, and ensures that individuals from different groups have equivalent diversity of recourse options (Detailed later in Sec. \ref{ProblemFormulationAndMehtodology}). To the best of our knowledge, this is the first work to introduce the notion of hybrid fairness for CF. We investigate to what extent considering hybrid fairness still allows individuals to achieve recourse, examine the trade-offs with respect to both individual- and group-level fairness, and analyze its impact on the quality and accessibility of recourse. The contributions of our work can be summarized as follows:
\begin{itemize}
    \item We model the problem of generating fair CFs as an optimization problem and propose a model-agnostic RL-based approach to optimize for fair CFs across different groups. We design customized reward functions to account for the diverse CF fairness requirements, such as equal effectiveness and equal choice of recourse, as well as CF quality metrics like validity and plausibility.

    \item We formulate individual, group, and hybrid fairness for CF generation and quantify the effectiveness of our proposed RL-based approach in guaranteeing fairness and achieving recourse for individuals and groups.

    \item We provide a comprehensive analysis that quantifies the impact of different fairness levels, including hybrid, individual, and group fairness, on CF quality. This analysis systematically compares our method against well-established baselines, highlighting the trade-offs between fairness enforcement and CF quality.
        

\end{itemize}

\section{Preliminaries on Counterfactual Explanations}\label{Preliminaries}

\subsection{Counterfactual Explanations}
A CF is defined relative to a specific data point, describing how changes in its features would alter model's prediction \cite{wachter2017counterfactual}, by constructing hypothetical scenarios that, if true, would have led to a different outcome (e.g., a favorable outcome). By comparing the actual outcome with corresponding CF explanations, stakeholders can evaluate the impact of specific factors and determine the influence of inputs on the outcome.

To generate CFs, one would optimize for diverse metrics such as validity, plausibility, similarity and minimality, and actionability \cite{guidotti2024counterfactual}. Validity refers to the ability to successfully change the ML model outcome compared to the original instance. Plausibility ensures that the CF feature values are realistic, i.e., they fall within the range of values observed in the dataset and are not flagged as outliers. Similarity and minimality (often referred to as proximity and sparsity) aim to keep the CF as close as possible to the original instance, minimizing both the number and magnitude of feature changes. Finally, actionability requires that all changes involve only features that a user can feasibly modify.

\subsection{Fair Counterfactual Explanations}
We consider fair CF generation at three levels, namely, individual, group and hybrid.

\paragraph*{\emph{Individual-level fairness}} Drawing upon the principle of \emph{treat similar individuals similarly}, individual-level fairness constrains that minor variations in input features should not result in significantly different CFs, as shown in Fig. \ref{fig:EF-Micro}. Fairness metrics employed for individual-level fair CFs often incorporate stability and robustness metrics \cite{artelt2021evaluating}, which aim to quantify the extent to which similar individuals (i.e., individuals with nearly identical input attributes) receive similar CF explanations (i.e., CF explanations that suggest changes in the same or similar attributes, and in comparable magnitude). 

\paragraph*{\emph{Group-level fairness}} Drawing upon the principle of \emph{fairness through unawareness}, group-level fairness refers to ensuring that different protected groups, which are defined as groups of individuals sharing a protected group (e.g. race, gender, nationality), have equal recourse opportunities to achieve favorable outcomes (i.e., the ability to achieve a recourse through CFs should not significantly vary in difficulty across protected groups). Fig. \ref{fig:EF-Macro} illustrates this concept by supposing the data subset in the shaded area, where the two protected groups are expected to receive similar CF adjustments (represented by the arrow) exemplifying how groups get a similar CF to cross the decision boundary. These adjustments aim to ensure a recourse by reducing disparities in achieving favorable outcomes, thereby preventing one group from facing disproportionately greater difficulty in obtaining recourse. Fig. \ref{fig:ECR} additionally demonstrates the equal opportunity concept, where both protected groups have equal choices to achieve recourse (a similar number of possible and valid recourses).

\paragraph*{\emph{Hybrid-level fairness}} In the context of CF generation, hybrid fairness aims to generate fair CFs that are actionable for individuals while addressing disparities across groups collectively. For example, in an educational setting, individual-level fairness ensures that two students with comparable academic performance and engagement receive similar CF suggestions (e.g., taking a specific preparatory course to qualify for a scholarship), while group-level fairness ensures that students from underrepresented or disadvantaged backgrounds are not consistently given harder or less attainable recommendations. Hybrid fairness ensures both personalized academic guidance and equitable opportunities across different demographic groups. Generating CFs under hybrid-level fairness is challenging, as it must balance individual and group fairness, which often have different focuses and conflicting objectives.

\section{Related Work}\label{RelatedWork}

\subsection{Generating Counterfactual Explanations}
The generation of CFs can be viewed as an optimization problem aimed at minimizing a cost function that encodes desirable CF explanation properties such as actionability \cite{mothilal2020dice,brughmans2024nice} (explained in more detail later in Sec. \ref{ProblemFormulationAndMehtodology}). The cost function may be model-dependent, i.e., it depends on the model's inner workings, such as gradients for differentiable models \cite{selvaraju2017grad,wang2020scout} or tree structures for tree-based models \cite{stepin2020generation,lucic2022focus}, or model-agnostic \cite{karimi2020model}. Numerous works focus on the CF generation problem investigating various methodologies such as mixed integer programming \cite{ustun2019actionable}, heuristic searches \cite{tolomei2017interpretable,martens2014explaining}, gradient-based approaches \cite{dhurandhar2018explanations,moore2019explaining}, and advanced meta-heuristics such as genetic algorithms \cite{sharma2019certifai,hashemi2020permuteattack}, GANS \cite{nemirovsky2022countergan} and RL \cite{numeroso2020explaining,nguyen2021counterfactual,ezzeddine2023sac}. These works aim to produce CFs aligned with standard optimization criteria (e.g., closeness to the original instance or actionability), but they do not explicitly incorporate fairness objectives.

More recently, research efforts began investigating fairness in relation to CFs \cite{artelt2022explain,asher2021fair,von2022fairness,kusner2017counterfactual,asher2022counterfactual}. The literature on fair CFs can be broadly distinguished into model-agnostic and model-specific approaches. While the former enforce fairness in the generation of CFs themselves, such as in authors in \cite{artelt2022explain} propose a model-agnostic method for generating group fair CFs while minimizing complexity of CFs across protected groups. The latter embeds CF-based fairness criteria into the training process to promote or evaluate fair model predictions. For instance, authors in \cite{asher2021fair} investigate and prove that CFs can hide model unfairness. The work in \cite{von2022fairness} defines fairness of recourse criteria at both individual and group levels that account for causal relationships between features. Similarly, \cite{russell2017worlds} explores fairness under multiple causal models, enabling fair decisions without requiring a precise causal specification. Research in \cite{rawal2020beyond} optimizes resource correctness, interpretability, and cost efficiency across populations and subpopulations to ensure fairness. Also, authors in \cite{han2023dualfair} enhance model fairness by retraining the model by optimizing for group fairness and counterfactual fairness by flipping the sensitive attribute without optimizing for fairness in the explainer.

Other research has focused on robustness, relating it to individual-fairness in CF generation \cite{artelt2024two,pawelczyk2022exploring,cheon2024feature,nguyen2024survey}, as unstable CFs can undermine reliability and lead to unfair treatment of specific groups \cite{artelt2021evaluating}. In \cite{artelt2024two}, the authors propose a methodology that clusters similar instances and generates a single CF for multiple instances, but does not account for fairness explicitly.

\subsection{Counterfactual Explanations for Auditing Fairness}

Beyond the works that incorporate fairness into CF generation, several studies have leveraged CFs to audit fairness in models by using the insights CFs provide to detect global discriminatory behavior, revealing whether similar individuals are treated differently in terms of explanations \cite{kavouras2024fairness,ley2023globe,kusner2017counterfactual,goethals2024precof,sharma2019certifai,cornacchia2023auditing,kuratomi2022measuring,wang2024advancing,garg2019counterfactual}. For example, in \cite{goethals2024precof}, the authors propose an approach to detect model bias by comparing sensitive and correlated attributes of CFs across protected groups while in \cite{cornacchia2023auditing}, the authors propose a bias detection framework that utilizes CFs to identify minimal-cost positive outcomes to detect proxies for sensitive features. In \cite{kuratomi2022measuring}, the authors define two metrics to quantify model fairness, burden, and normalized accuracy-weighted burden, which integrate accuracy and counterfactual-based fairness measures using CFs. Also, authors in \cite{ma2022learning} propose graph counterfactual fairness to evaluate fairness from a causal perspective, comparing each individual’s original prediction to its CF. \cite{coston2020counterfactual} investigates connections between standard fairness metrics and their CF counterparts. Authors in \cite{ley2023globe} introduce \emph{GLOBE-CE}, a method to generate summaries for global CF-based explanations that are used to detect fairness. Also, \cite{kavouras2024fairness} presents a framework, referred to as \emph{FACTS}, to audit fairness through CF resources at both individual and subgroup levels, addressing individual and group-level fairness through detecting statically embedded unfairness by applying item set mining algorithms on the data and computing the proposed metrics. Similarly, \cite{fragkathoulas2024fgce} introduces a graph-based framework for generating group CF with shared properties of explanations with the goal of audit model fairness. For a more in-depth analysis on the difference between our work and works that audit for fairness with CF, we highlight in Tab.~\ref{tab:comparison}  the main distinctions between our work and the most relevant approaches, namely GLOBE-CE \cite{ley2023globe}, FACTS \cite{kavouras2024fairness}, and FGCE \cite{fragkathoulas2024fgce}.
\begin{table}[htbp]
\centering
\small
\caption{Comparison of GLOBE-CE \cite{ley2023globe}, FACTS \cite{kavouras2024fairness},FGCE \cite{fragkathoulas2024fgce} and our RL-Based CF Generation Approaches}
\begin{tabular}{p{2cm}|p{2cm}|p{2cm}|p{2cm}}
\textbf{GLOBE-CE} & \textbf{FACTS} & \textbf{FGCE} & \textbf{RL-Based} \\ \hline

\textbf{Summarize} global CFs to audit model fairness 
& \textbf{Audit fairness} by analyzing sub-group recourse metrics
& \textbf{Audit fairness} by generating group CFs for shared recourse properties
& \textbf{Enhance fairness} in CF explanations (EE, ECR and cost)  \\ \hline

Evaluates global patterns (no explicit sub-group focus)
& Statically compares fairness among item-mined sub-groups
& Groups individuals by CF similarity, enabling sub-group model fairness analysis
& Dynamically optimizes individual, group, and hybrid CF recourse fairness \\ \hline

Employs translation-based techniques to identify global directions
& Uses item mining (with binning for continuous features) for feature exploration
& Explores weakly connected components in a feasibility graph to identify similar CFs
& Balances exploration and exploitation and navigates continuous and categorical feature spaces \\ 
\end{tabular}
\label{tab:comparison}
\end{table}

In this work, we focus on the generation of fair CFs, in contrast to prior studies that primarily employ CFs as tools for auditing fairness. We propose a model-agnostic RL–based framework that explicitly incorporates novel fairness constraints into the CF generation process of a fair model, thereby enabling the systematic treatment of individual, group, and hybrid fairness on the explanation level. Moreover, we contribute to the formulation of hybrid fairness, as it unifies individual- and group-level considerations, which is, to the best of our knowledge, has never been addressed before. Additionally, our method generates, by design, a diverse collection of CFs from which individuals across different groups can equitably select, ensuring broad and fair access to viable recourse options while simultaneously optimizing well-established CF quality metrics such as actionability, sparsity, and proximity. We also extend the fairness definitions introduced by \cite{kavouras2024fairness}, shifting the focus from auditing model fairness to positioning fair CF generation as a primary objective, and design customizable RL reward functions accordingly. Finally, by embedding fairness constraints directly into the generation process, our approach moves beyond cost-centered objectives and provides a principled foundation for quantifying the trade-offs between fairness and other conventional evaluation metrics. 

\section{Problem Formulation and Methodology}\label{ProblemFormulationAndMehtodology}
In this section, we first present the foundational components of our methodology, including the data and model representation. We then introduce the fairness metrics employed in our analysis. Finally, we then detail the RL-based methodology designed to optimize for these metrics.

\subsection{Data, Model and Counterfactual Representation}
We follow the definitions outlined in \cite{kavouras2024fairness} regarding the data, model, and input space representation. We also illustrate how a CF is represented as an action in the feature space and formally define what constitutes a group.

The feature space, denoted by \( \mathcal{X} \), is defined as the Cartesian product of \( n \) individual feature spaces, that is, $\mathcal{X} = \mathcal{X}_1 \times \mathcal{X}_2 \times \cdots \times \mathcal{X}_n$. Among these, the feature space \( \mathcal{X}_n \) represents a sensitive or protected attribute that takes binary values \( 0 \) or \( 1 \). Each individual is represented by a single instance \( x \in \mathcal{D} \) where \( \mathcal{D} \) is the training dataset. We consider a binary classifier \( h: \mathcal{X} \to \{0, 1\} \), which maps each instance \( x \in \mathcal{D} \) to a label in the set \( \{0, 1\} \). The label \( 1 \) denotes a favorable outcome, while the label \( 0 \) signifies an unfavorable outcome. The classifier \( h \) operates over the feature space \( \mathcal{X} \), providing a decision for each instance based on its attributes.

A \emph{group} $G$ is a subset of the dataset \(D\) and it consists of distinct individuals who each possess a sensitive or protected attribute \(\mathcal{X}_n\). Individuals of the same group share similar attributes and are positioned similarly within the feature space. Considering a sensitive attribute, we define two sub-groups:
\[
  G_0 = \{\, x \in D : X_n = 0 \}
  \quad\text{and}\quad
  G_1 = \{\, x \in D : X_n = 1 \}.
\]
where \( G_0 \) consists of individual instances with the sensitive attribute \( \mathcal{X}_n = 0 \), and \( G_1 \) consists of individuals with \( \mathcal{X}_n = 1 \). Our analysis focuses on the \emph{affected dataset} \( D_{affected} \subseteq \mathcal{X} \), which consists of instances from \( \mathcal{X} \), including individuals who have been classified unfavorably, i.e., for whom \( h(x) = 0 \).

An \emph{action} represents a CF as a transformation that modifies feature values and is denoted by \( a \in \mathcal{A} \), where \( \mathcal{A} = \{a_1, a_2, \ldots, a_i\} \) is the set of possible actions, where each action defines a way to alter the features of an instance \( x \in D \). We call $x'$ the \emph{CF instance} of $x$ under the action $a$ ($x' = a(x)$) if \( a \) provides \emph{recourse} to \( x \) and change the classifier’s decision from unfavorable to favorable, that is, if \( h(x) = 0 \) and \( h(x') = 1 \).
For example, consider a loan approval classifier \( h \) that predicts whether a loan application is approved (\( h(x) = 1 \)) or denied (\( h(x) = 0 \)). Let an instance \( x \) represent a loan application with a credit score of 580 and an annual income of 30k. Suppose the classifier outputs \( h(x) = 0 \), indicating the loan application is denied. An action \( a \in \mathcal{A} \) can be to increase the credit score by 40 and the income by 5k. Applying this action transforms the instance \( x \) into a new instance \( a(x) \), which has a credit score of 620 and an income of 35k. If the classifier now outputs \( h(a(x)) = 1 \), the action \( a \) is considered to provide \emph{recourse} to the applicant, as it changes the decision from unfavorable to favorable.

\subsection{Counterfactual Fairness Metrics}
To achieve these fairness objectives, we need to optimize for two terms, namely, \emph{equal effectiveness (EE)} and \emph{equal choice of recourse (ECR)}. Following the work of \cite{kavouras2024fairness}, we aim to define \emph{EE} and \emph{ECR}, which were primarily used as metrics to audit the fairness of ML models.

We start with \emph{effectiveness}, as a metric to evaluate the fairness of a single action $a$ (one CF) on a subset of the data. The \emph{effectiveness} (also referred to as \emph{coverage}) of an action \( a \) quantifies its utility across a group of individuals. Specifically, it measures the proportion of individuals in an affected group \( G \subseteq D \) who would receive a favorable outcome if the action \( a \) were employed for all of them. This metric reflects how broadly applicable or beneficial a given action is within the group. Formally, the effectiveness of action \( a \) for group \( G \) is:
\begin{equation}
    \text{eff}(a, G) = \frac{1}{|G|} \left | \{x \in G \mid h(a(x)) = 1\} \right|.
    \label{eq:action_effectiveness}
\end{equation}

To evaluate individual and group fairness of a set of actions $\mathcal{A}$ for a group $G$. We consider individual-effectiveness (accounting for individual fairness) and group-effectiveness (accounting for group fairness), respectively.

\emph{Individual-Effectiveness:}
From a individual perspective, individuals are examined independently of the group they belong to. Similar individuals select an action from $\mathcal{A}$ that is most advantageous for them, as shown in Fig. \ref{fig:EF-Micro}. Individual-effectiveness measures the proportion of individuals who can achieve recourse by choosing one action $a$ from $\mathcal{A}$ that, when employed, achieves their desired outcome at the least cost:
\begin{equation}
    \text{aeff}_{\mu}(\mathcal{A}, G) = \frac{1}{|G|} \left| \{x \in G \mid \exists a \in \mathcal{A}, h(a(x)) = 1\} \right|.
    \label{eq:micro_effectiveness}
\end{equation}

\emph{Group-Effectiveness:}
From a group perspective, a group $G$ is treated collectively with a single action applied to all individuals, meaning examining the overall outcome for the group as a whole, as shown in Fig. \ref{fig:EF-Macro}. The group-effectiveness of a set of actions $\mathcal{A}$ for a group $G$ is defined as the maximum proportion of individuals in $G$ who can achieve recourse through the same action:
\begin{equation}
    \text{aeff}_{M}(\mathcal{A}, G) = \max_{a \in \mathcal{A}} \frac{1}{|G|} \left| \{x \in G \mid h(a(x)) = 1\} \right|.
    \label{eq:macro_effectiveness}
\end{equation}

Maximizing group-effectiveness across different CFs ensures that the framework can identify actions that are collectively beneficial across the group, minimizing disparities and ensuring fair outcomes for all members.

\subsection{Problem Formulation and Objectives}
Given a fair binary classifier \( h: \mathcal{X} \rightarrow \{0, 1\} \), where \( 0 \) indicates a unfavorable outcome. An affected dataset $D_{affected}$ consists of individuals who received unfavorable outcomes ($h(a(x) = 0$). $D_{affected}$ consists of protected groups \( G = \{G_1, G_2\} \), which partition $D_{affected}$, based on sensitive attributes (e.g., ethnicity). We ensure fair training for the models, as our objectives aim to avoid fairwashing through the explanations we provide.  The objective is to generate a set of CF \( \mathcal{A} \) that enhances fairness at both individual and group levels. Specifically \( \mathcal{A} \) should:
\begin{enumerate}
    \item Maximize the proportion of individuals that can find effective recourse by applying actions from \( \mathcal{A} \).
    \item Guarantee \emph{Fair Opportunity}, so that all individuals and groups should have a fair chance to improve their outcomes through the provided CFs.
    \item Identify valid, actionable, plausible CFs and similar to the original instance and that offer practical steps that individuals across groups can realistically implement.
    \item Ensure that every group have access to a diverse set of CFs, preventing any unfairness in terms of diversity, and is not limited to only one hard-to-implement action.
\end{enumerate}

\emph{Equal Effectiveness (EE)} requires that similar proportions of individuals from the two protected groups G0 and G1 achieve recourse through the provided set of actions. This ensures that no group is disproportionately disadvantaged in terms of the effectiveness of the CF. Given the individual- and group-effectiveness metrics $\text{aeff}_{\mu}(\mathcal{A}, G)$ and $\text{aeff}_{M}(\mathcal{A}, G)$ for groups $G_0$ and $G_1$, EE is achieved when:
    \begin{equation}
        |\text{aeff}_{\mu}(\mathcal{A}, G_0) - \text{aeff}_{\mu}(\mathcal{A}, G_1)| \leq \epsilon 
         \label{eq:equal_effectiveness_micro}
    \end{equation}
    \begin{equation}
        \text{and} \quad |\text{aeff}_{M}(\mathcal{A}, G_0) - \text{aeff}_{M}(\mathcal{A}, G_1)| \leq \epsilon
        \label{eq:equal_effectiveness_macro}
    \end{equation}
where $\epsilon$ is a small, predefined threshold quantifying disparities.

\emph{Equal Choice for Recourse (ECR)} ensures that individuals from different protected groups have equal opportunities (or number of actions) to achieve recourse from \( \mathcal{A} \). This metric addresses disparities in the diversity and feasibility of actionable explanations provided to different groups. Let $\mathcal{A}_G$ represent the subset of actions from $\mathcal{A}$ that are effective for group $G$, an action \( a \in \mathcal{A} \) is considered effective if it achieves recourse for at least a proportion of individuals ($\phi$) from group $G$. ECR is achieved when the two groups $G0$ and $G1$ have the same number of valid actions:
    \begin{equation}
        |\mathcal{A}_{G_0}| = |\mathcal{A}_{G_1}| \quad
        \label{eq:equal_choice_for_recourse}
    \end{equation}

\subsection{RL-based Approach For Generating Fair Counterfactuals}
We design an RL agent with a Soft Actor-Critic (SAC) algorithm to serve as a CF generator. The choice of RL, specifically SAC (detailed in Appendix), is due to its sequential and exploratory nature in feature space, which makes it suitable for the sequential nature of achieving a recourse. Specifically, the entropy regularization step ensures stochastic policies, making it well-suited for high-dimensional action spaces. Our objective is not to identify a single static CF or recourse path, but rather to generate diverse, realistic, and feasible CFs that provide recourse to data points to reach a desired target. An RL sequential environment supports our aim by enabling the model to condition on previous actions, such that each modification of an actionable feature (for example, education, lifestyle, or employment) alters the feasible space of subsequent actions. In contrast, static optimizers treat all changes as independent, which can lead to unrealistic one-shot interventions. Furthermore, RL frameworks balance short-term and long-term trade-offs, as sequential decision-making naturally incorporates cumulative costs or constraints, such as effort, time, or limited resources, which are challenging to encode in static solvers. By learning a policy rather than solving a fixed optimization problem for each instance, the SAC sequential agent can explore multiple feasible paths and generate diverse recourse sets, rather than a single deterministic solution. 

This makes the RL promising as a robust value estimation and scalable optimization to support the process of CF generation. As shown in Sec. \ref{RelatedWork}, RL is a suitable method for generating CF and offers a flexible and robust framework that addresses several limitations of traditional optimization approaches. Unlike gradient-based, manifold-based, or convex optimization methods, which often treat CF generation as a static optimization problem, RL frames it as a sequential decision-making process. RL also enables effective exploration of the CF space, which is especially important in non-differentiable or discrete domains.
The SAC takes as input the classifier $h$ and the affected dataset \( D\_{affected} \) and provides as output a set of CFs that represent $\mathcal{A}$. We follow the Markov Decision Process (MDP) problem formulation, which involves defining state and action spaces and customized reward functions. Since SAC is a model-free RL method, it does not require defining a transition function. In the following subsection, we detail the state and action representation along with the reward functions. Specifically, we design 3 reward functions that incorporate the EE and ECR objectives $R_{EE}$ (at both, individual and group levels) and $R_{ECR}$ (group) and $R_{EE-ECR}$ (hybrid), respectively, to ensure that the CFs adhere to the fairness metrics in Eq. \ref{eq:equal_effectiveness_micro}, \ref{eq:equal_effectiveness_macro} and \ref{eq:equal_choice_for_recourse}, respectively.

\subsubsection{State Representation}
\begin{figure}[htp]
    \centering
        \includegraphics[width=0.5\textwidth]{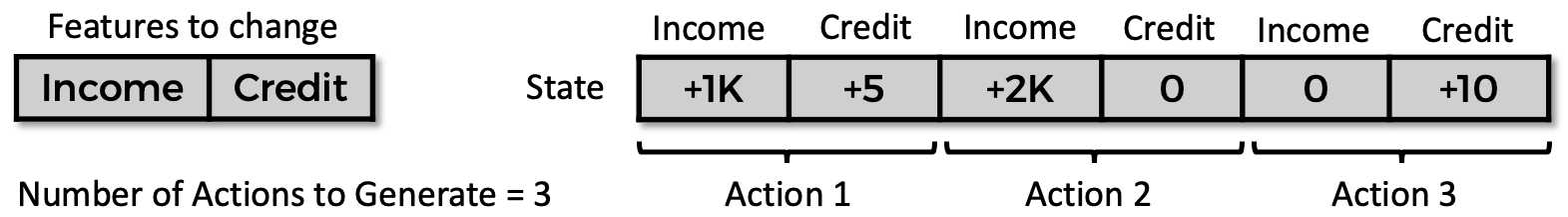}
        \caption[RL State Representation]{State representation, given the set of features to change and the maximum number of actions to generate.} 
        \label{fig:state_representation}
\end{figure}

Suppose $n$ is the number of desired actions $a$ (CF) to be generated to form \( \mathcal{A} \) ($n=|\mathcal{A}|$), \( \mathcal{F}_\text{act} \) the set of actionable features (mutable features), and $l = |F_\text{act}|$ (ensuring diversity and actionability). The state $s$ is represented as an $N$-dimensional vector where $N = n \times l$ that corresponds to a concatenation of features representing the amount of change to the value of each actionable feature. Fig. \ref{fig:state_representation} shows an example of the state vector where \( \mathcal{F}_\text{act} = \{Income, Credit \} \) and the number of actions to generate is equal to 3, then the state will have a dimension of 6.
   
\subsubsection{Action Representation} 
\begin{figure}[htbp]
        \centering
        \includegraphics[width=0.5\textwidth]{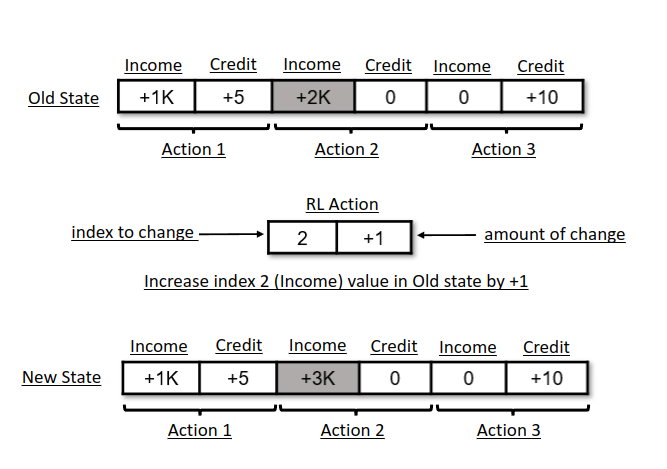}
        \caption[RL Action Representation]{RL action representation, and the change in state after applying the action.}
        \label{fig:action_representation}
        \vspace{-0.5cm}
\end{figure}

An action is represented as a 2-dimensional vector $a = \{a_1, a_2\}$, where $a_1$ indicates the index of the feature to be modified within the state, at each time step, and $a_2$ indicates the amount of change to the current value of the corresponding feature. This representation enables the RL agent to learn not only the optimal amount of change but also which feature to modify at each step, thereby accounting for minimality and similarity. Note that we apply normalization and denormalization to the feature space of actions to achieve a consistent scale, as it helps improve the stability and efficiency of the learning process and prevents features with larger scales from dominating the optimization. The proposed actions feature values are bounded by each feature range to ensure plausibility when aligned with similarity measures. Fig. \ref{fig:action_representation} shows an illustrative example of RL action that suggests changing the feature at index 2 (feature \emph{income}) with an amount of +1k, resulting in a change in the feature's value from 2k to 3k.

\subsubsection{Reward Function:} The RL reward function incorporates the fairness metrics to be optimized, EE and ECR, and to the Gower distance, to ensure group fairness and similarity, respectively. We define $R_{EE}$, $R_{ECR}$, and $R_{EE-ECR}$. Specifically, $R_{EE}$ is separately used when we aim to optimize only for EE, while $R_{ECR}$ is separately used when we aim to optimize for ECR, while $R_{EE-ECR}$ is used when we aim to optimize for EE and ECR simultaneously. 

The reward function for the Individual- and Group- EE case consists of the following terms (Eq. \ref{eq:equal_effectiveness_micro} and Eq. \ref{eq:equal_effectiveness_macro}), presented in Eq. \ref{eq:reward_function1}:
         \begin{itemize}
            \item \emph{Average success rate (ASR)}: Average proportion of individuals who achieve recourse using any of the proposed actions for the two protected groups: $(SR_1 + SR_2)/2$, where $SR_i$ represents the success rate (SR) of group $i$.
            \item \emph{Proportions difference (PD)}: Absolute difference between the proportions of individuals across protected groups who can achieve recourse using any of the proposed actions: $|SR_1  - SR_2|$.
            
            \item \emph{Similarity (Sim)}: The average Gower distance of points that were able to achieve recourse through an action $a$ from $\mathcal{A}$: $Gower (x, h(a(x))$.
            
            \item \emph{Active actions (Act):} number of active actions (an action that could be used to achieve recourse), it is used to ensure a diverse set of non-zero actions: $|a \in \mathcal{A}, \quad eff (a, G) \geq \alpha|$.
            
            \item $\mathbf{C_{0,1,2,3}}$: regularization terms allowing to balance trade-offs according to desired priorities. 
         \end{itemize}
    \begin{align}
        R_{EE}(s_{t-1}, a, s_t) &= ASR_{s_t} \times C_{0} + Act_{s_t} \times C_{1} \nonumber \\
        &\quad - PD_{s_t} \times C_{2}  - Sim_{s_t} \times C_{3} 
        \label{eq:reward_function1}
    \end{align}

To ensure ECR (Eq. \ref{eq:equal_choice_for_recourse}), we customize the reward function to encourage diverse CF choices across protected groups by maximizing the number of actions available to each group. Additionally, we include terms that maintain fairness between the groups, ensuring equal potential for recourse, for example, by considering the number of active actions. Specifically, the reward function (Eq. \ref{eq:reward_function2}) is:

         \begin{align}
            R_{ECR}(s_{t-1}, a, s_t) &= A_{0s_t} \times C_{0} + A_{1s_t} \times C_{1} - AD_{s_t} \times C_{2} \nonumber \\
            &\quad - Act_{s_t} \times C_{3} - Sim_{s_t} \times C_{4} 
            \label{eq:reward_function2}
        \end{align}
        \begin{itemize}
            \item \emph{$A_0$ and $A_1$:} number of actions for the two protected subgroups $G_0$ and $G_1$, respectively (e.g., males and females that can be used to achieve recourse).
            
            \item \emph{Action Difference (AD):} number of actions difference represents the absolute difference between the number of actions available for each protected subgroup to achieve recourse: $|A_0 - A_1|$.
            
            \item \emph{Similarity (Sim)}: The average Gower distance of points that were able to achieve recourse through an action $a$ from $\mathcal{A}$: $Gower (x, h(a(x))$.
            
            \item \emph{Active actions (Act):} number of active actions (an action that could be used to achieve recourse), it is used to ensure a diverse set of non-zero actions: $|a \in \mathcal{A}, \quad eff (a, G) \geq \alpha|$.
            
            \item $\mathbf{C_{0,1,2,3,4}}$: regularization terms allowing to balance trade-offs according to desired priorities \footnote{The regularization parameters can be adjusted to accommodate user constraints and preferences. In this work, equal weight is assigned.}. 
        \end{itemize}

    For the case of the hybrid approach, which optimizes for the hybrid fairness, we specify the reward function $R_{EE-ECR}$ that represents a combination of $R_{EE}$ and $R_{ECR}$:
        \begin{align}
            R_{EE\text{-}ECR}(s_{t-1}, a, s_t) &= R_{EE}(s_{t-1}, a, s_t) \nonumber \\
            &\quad + R_{ECR}(s_{t-1}, a, s_t)
            \label{eq:reward_function3}
        \end{align}

\subsubsection{Stopping criteria} The episode-stopping condition that specifies the reach of the RL goal is met when the following criteria are satisfied:
    \begin{itemize}
        \item For EE: $ ASR >= target\_success\_rate$ and $PD <= target\_proportion\_difference$.
        \item For ECR: $A_0 >= target\_group0\_nb\_actions$ and $A_1 >= target\_group1\_nb\_actions$ and $AD <= target\_nb\_actions\_difference$.
    \end{itemize}

\subsection{Counterfactual Evaluation Metrics}

We design our approach to satisfy several key criteria shown in the rewards function. Actionability is ensured by allowing users to specify the features they are willing to change. We achieve validity through direct optimization, while plausibility is enforced by constraining the search space of our optimizer. Finally, we optimize for similarity using Gower distance. A CF $a$ is evaluated based on several key metrics:
\begin{itemize}
    \item \emph{Validity.} A CF is considered \emph{valid} if it successfully changes the classifier’s prediction compared to the original instance. Formally, for a given instance \( x \in D \) with \( h(x) = 0 \), a CF \( a(x) \) is valid if \( h(a(x)) = 1 \). For a dataset $D$, the validity is the proportion of individuals for whom the generated CF achieves the desired outcome:
    \[
    \text{Validity} = \frac{1}{N} \sum_{i=1}^{N} \mathbf{1}\left[h(a(x_i)) = y_{\text{target}}\right]
    \]

    \item \emph{Plausibility.} A CF is \emph{plausible} if its feature values lie within the data distribution, i.e., they do not correspond to outliers. This is needed to ensure that the CF represents a realistic instance. One of the robust methods to compute plausibility is based on the reconstruction error from a denoising autoencoder (following~\cite{van2021interpretable,nemirovsky2022countergan}):
    \[
    \text{Plausibility} = \|a(x_i) - \hat{a(x_i)}\|^2
    \]
    where $\hat{a(x_i)}$ is the autoencoder reconstruction of the CF $a(x_i)$. Lower the reconstruction error, meaning the CFs fit the training data distribution and are more realistic.

    \item \emph{Similarity and Minimality.} A CF should be similar to the original instance. \emph{Similarity (proximity)} refers to how close the CF is to the original instance in feature space, often measured using a distance function \( d(x, a(x)) \) such as Gower distance. \emph{Minimality} (or sparsity) focuses on minimizing the number of features changed:
    \[
    \text{Similarity} = \frac{1}{N} \sum_{i=1}^{N} \text{Gower}(x_i, a(x_i))
    \]
     \[
    \text{Minimality} = \frac{1}{N} \sum_{i=1}^{N} \|x_i - a(x_i)\|_0
    \]

    \item \emph{Actionability.} A CF is \emph{actionable} if it only modifies features that the individual can feasibly change. Let \( \mathcal{F}_\text{act} \subseteq \mathcal{F} \) denote the set of actionable features. Then, a CF \( a(x) \) is actionable if and only if \( \text{supp}(x - a(x)) \subseteq \mathcal{F}_\text{act} \), where \( \text{supp}(\cdot) \) denotes the set of features with changes.
\end{itemize}

\section{Evaluation Setup}\label{ExperimentalSettings}
This section outlines our experimental setup, including the datasets utilized for analysis and the RL configuration employed to evaluate the proposed methodology. Our framework's implementation is available online.\footnote{\url{https://github.com/ObaidaAmmar/IGH-Fair-CFE}}

\subsection{Experimental Scenarios and Baselines}
We consider four fairness scenarios, as follows:

\subsubsection{Individual - Equal Effectiveness (Individual-EE)} In this scenario, each individual, regardless of their protected attribute value, chooses the recourse action from the list of generated actions that best addresses their unique situation. Each individual chooses the action that has the lowest Gower distance (lowest recourse cost) compared to the original instance. We run the RL considering Eq. \ref{eq:micro_effectiveness} and \ref{eq:equal_effectiveness_micro} and optimize for $RR_{EE}$.

\subsubsection{Group - Equal Effectiveness (Group-EE)} In this scenario, the aim is to identify a single widely applicable action that achieves recourse for a least effectiveness of $eff(a, G)=75\%$) of individuals across both protected groups. This approach prioritizes broad effectiveness while aiming to provide recourse to all individuals within a group, where we run the RL on Eq. \ref{eq:macro_effectiveness} and \ref{eq:equal_effectiveness_macro} and optimize for $RR_{EE}$.

\subsubsection{Group Equal Choice for Recourse (Group-ECR)} This scenario represents a group ECR. It identifies a set of actions that offer ECR to both protected subgroups. An action is considered effective if it results in recourse for more than $\phi=60\%$ \footnote{This represents the minimum boundary, however, the SR is optimized by the reward function, which aims to maximize the $\phi$.}of individuals in that affected dataset, by ensuring Eq. \ref{eq:equal_choice_for_recourse} and optimizing for $RR_{ECR}$.

\subsubsection{Hybrid Individual-Equal Effectiveness and Group-Equal Choice for Recourse (Hybrid-EE-ECR)} This scenario represents the hybrid fairness, which balances between individual and group fairness. It explores feature space that allows for individual fairness while ensuring that a significant proportion of individuals from each protected group can achieve recourse through a set of collectively effective actions, based on Eq. \ref{eq:micro_effectiveness}, \ref{eq:equal_effectiveness_micro} and \ref{eq:equal_choice_for_recourse} simultaneously and optimize for $RR_{EE-ECR}$.

\subsubsection{Clustering} In our experiments, we cluster the affected dataset into 3\footnote{We fix the number of clusters to three for simplicity and consistency across datasets, also to reflects a balance between granularity and interpretability: too few clusters would fail to capture heterogeneity, while too many would lead to overly fragmented groups, making analysis and comparison less meaningful.} clusters using the k-means algorithm \cite{lloyd1982least,macqueen1967some}, creating groups of individuals with similar feature values close in space regardless of the protected attributes. This clustering step allows us to explore fair CF both globally across the entire dataset and within each chosen cluster. We evaluate fairness metrics on the whole affected dataset (we refer to this by \emph{Whole} in the rest of the paper) as well as on each cluster separately (C1-C3). For each individual, we select the best valid CF that has the lowest Gower distance, and treat it as the final CF. Additionally, we report the clustering-based approach, where each individual is assigned the CFs generated for the cluster to which they belong.

In our experiments, we aim to generate a maximum of 5 actions for EE, ECR and EE-ECR. We ensure that most individuals can successfully achieve recourse and that the PD across protected groups does not exceed 10\% (the upper bound is 100\%), to maintain fairness in the distribution of successful recourse outcomes. For ECR, we impose constraints to ensure that the \emph{number of actions} for each protected group is at least 1 to guarantee sufficient recourse options, and that the \emph{number of actions difference} between the two protected groups is 0, to ensure protected groups receive ECR.

\paragraph{Baselines} To evaluate the effectiveness of our RL-based CF generation methods, we compare them against 3 widely-adopted CF methods such: Nearest Instance Counterfactual Explanations (NiCE) \cite{brughmans2024nice}, Diverse Counterfactual Explanations (DiCE) \cite{mothilal2020explaining}, and prototype-guided CFs \cite{van2021interpretable}. Each method is configured with hyperparameters optimized to maximize the quality of CFs, balancing proximity, sparsity, and plausibility.
\paragraph{NiCE} generates CFs by iteratively modifying the original instance by replacing feature values with those from the nearest unlike neighbor. We configure NiCE with the distance metric as Heterogeneous Euclidean-Overlap Metric to handle mixed feature types with min-max scaling for numerical features. The optimization objective is set to proximity and plausibility constraints.

\paragraph{DiCE} Generates diverse CFs by optimizing validity, proximity, and diversity. We set the Generation method to random sampling to generate 5 CFs per instance, ensuring diversity. The feature constraints are set to only mutable features, which are chosen as the ones used in our RL method.
    
\paragraph{Prototype-guided} Generate CFs by moving instances toward class prototypes. The prototype search method is the KD-tree acceleration with Loss weighting. The high prototype weight is set to 10 to ensure plausibility. The search process is optimized for 200 iterations and 15 perturbation steps. Along with Feature bounds derived from training data to avoid unrealistic values.

\subsection{Datasets Description}
We evaluate our approach using 3 widely-adopted datasets in fairness evaluation: Adult, SSL, and Alzheimer, each selected for their relevance and the presence of actionable features that enable meaningful interventions. These datasets collectively span diverse domains: economic, judicial, and medical, each with distinct fairness challenges and actionable recourses. Their inclusion ensures that our approach is robust, generalizable, and grounded in real-world impact.

\emph{Adult \cite{adult_2}:} Its use is motivated by the societal importance of fair decision-making in employment and financial contexts. Consists of 48842 data points. The classifier predicts whether an individual's income exceeds 50k/year. The dataset consists of 15 features representing demographic and socioeconomic factors. The protected attribute is Race, and the actionable features are continuous: capital-gain, hours-per-week, and categorical: educational-num. Actionable features such as capital-gain, hours-per-week, and educational-num represent realistic features that individuals or institutions might adjust to improve outcomes.
\emph{SSL} Consists of 398,582 data points related to criminal justice. The classifier predicts the SSL score based on 10 features, including arrests, assaults, shootings, affiliation, and criminal activity. The protected attribute is Gender, and the actionable features are age at latest arrest (categorical) and trend in criminal activity (continuous). The actionable features can be influenced through policy or rehabilitation programs in justice-related applications.

\emph{Alzheimer's \cite{rabie_el_kharoua_2024}:} Consists of 2149 data points, designed to predict whether an individual will develop Alzheimer. It consists of 21 demographic, lifestyle, and health-related features. The protected attribute is Gender, and the actionable features are continuous: functional assessment, ADL, and MMSE. The actionable features represent clinical and lifestyle indicators that can be monitored or improved through intervention.

\subsection{Models training}
We evaluate the performance of the classifier, focusing on standard metrics (accuracy, precision, recall, and F1-score) and fairness measures (Equalized Odds (EQ) and Demographic Parity differences (DP)) to assess model fairness across protected groups. We also employ in-processing fairness training using the Exponentiated Gradient to enhance the fairness of the underlying ML model prior to running our experiments. Each dataset is split into 80\% for training and 20\% for testing. We train different ML models and choose the one that performs best in terms of fairness metrics and predictive performance. More specifically, for \emph{Alzheimer}, we train a Random Forest classifier, with max depth of 15, max features of 'sqrt', min samples leaf of 2, min samples split of 5, and 200 estimators. For \emph{Adult}, we train a Catboost classifier with default configuration. For \emph{SSL}, we train an XGBoost classifier with a subsample of 0.9, number estimators of 200, a max depth of 3, and a learning rate of 0.1.

The ML models trained on the various datasets achieved high predictive performance and acceptable fairness, as indicated by EQ and PD values below 10\%. Given that these metrics have an upper bound of 100\%, a level under 10\% is considered satisfactory, as it supports that our generated CFs do not contribute to fairwashing. For \emph{Alzheimer}, the model achieves an accuracy of 95.8\%, a precision, recall, and f1-score of 96\%. Fairness evaluation shows an EO difference of 2\% and a DP of 8\% between the two protected groups. For the \emph{Adult}, the model achieves an accuracy of 87.3\% and a precision, recall, and f1-score of 87\%. Fairness evaluation revealed an EO difference of 4\% and a DP difference of 6\%. For the \emph{SSL}, the model achieves an accuracy of 90\%, precision of 95\%, recall of 90\% and f1-score of 92\%. Fairness evaluation shows an EO difference of 2\%, and a DP of 0.9\%.

\section{Experimental Results}
In this section, we discuss experimental results in terms of the fairness of CFs generated across the various scenarios and the quality of the CFs.

\begin{table*}[ht]
\centering
\caption{SR across Whole and clusters C1–C3. Values are shown as [G1, G2] with the absolute (\textbf{PD}) in parentheses.}
\begin{scriptsize}
\begin{tabular}{llcccc}
\toprule
\textbf{Dataset} & \textbf{Scenario} & \textbf{Whole} & \textbf{C1} & \textbf{C2} & \textbf{C3} \\
\midrule
\multirow{4}{*}{\textbf{Alzheimer}} 
  & Individual-EE     & [87, 83] (\textbf{4}) & [84, 88] (\textbf{4}) & [91, 95] (\textbf{4}) & [100, 100] (\textbf{0}) \\
  & Group-EE          & [76, 79] (\textbf{3}) & [88, 91] (\textbf{3}) & [94, 100] (\textbf{6}) & [89, 84] (\textbf{5}) \\
  & Group-ECR         & [61.6, 61.1] (\textbf{0.5}) & [64, 62.5] (\textbf{1.5}) & [79.4, 90.5] (\textbf{11.1}) & [77.8, 68.4] (\textbf{9.4}) \\
  & Hybrid-EE-ECR     & [85, 86] (\textbf{1}) & [92, 100] (\textbf{8}) & [100, 100] (\textbf{0}) & [96, 95] (\textbf{1}) \\
\midrule
\multirow{4}{*}{\textbf{Adult}} 
  & Individual-EE     & [100, 100] (\textbf{0}) & [97, 98] (\textbf{1}) & [98, 99] (\textbf{1}) & [98, 99] (\textbf{1}) \\
  & Group-EE          & [100, 100] (\textbf{0}) & [100, 100] (\textbf{0}) & [99, 100] (\textbf{1}) & [99, 100] (\textbf{1}) \\
  & Group-ECR         & [97,5, 98.5] (\textbf{1}) & [97.3, 98.2] (\textbf{0.9}) & [96.8, 98.2] (\textbf{1.4}) & [97.8, 98.9] (\textbf{1.1}) \\
  & Hybrid-EE-ECR     & [99, 100] (\textbf{1}) & [98, 99] (\textbf{1}) & [98, 99] (\textbf{1}) & [98, 99] (\textbf{1}) \\
\midrule
\multirow{4}{*}{\textbf{SSL}} 
  & Individual-EE     & [98, 94] (\textbf{4}) & [96, 95] (\textbf{1}) & [95, 93] (\textbf{2}) & [100, 99] (\textbf{1}) \\
  & Group-EE          & [98, 94] (\textbf{4}) & [96, 94] (\textbf{2}) & [99, 98] (\textbf{1}) & [100, 98] (\textbf{2}) \\
  & Group-ECR         & [98, 92] (\textbf{6}) & [95, 92] (\textbf{3}) & [95, 94] (\textbf{1}) & [95, 93] (\textbf{2}) \\
  & Hybrid-EE-ECR     & [98, 92] (\textbf{6}) & [97, 96] (\textbf{1}) & [92, 90] (\textbf{2}) & [100, 100] (\textbf{0}) \\
\bottomrule
\end{tabular}
\end{scriptsize}
\label{tab:sr_groups_with_pd}
\end{table*}

\subsection{Success Rates and Proportion Difference}
Tab. \ref{tab:sr_groups_with_pd} reports the SR achieved by our approach across \emph{Whole}, the clusters C1, C2 and C3, and across the protected G1 and G2, considering the various fairness scenarios on 3 datasets. For each cell, we report the SR for both groups in the format \emph{[G1, G2]}, followed by the \emph{PD} highlighted in bold.

\paragraph*{Success Rate Across Fairness Scenarios}
Results show that the SRs achieved by our approach across diverse scenarios and datasets are relatively high, indicating its ability to generate valid CFs (i.e., CFs that enable individuals to perform recourse) for \emph{Whole} and across a variety of clusters (subgroups). Starting with the \emph{Adult}, our approach consistently achieves high SR, ranging between [97, 98] and [100, 100]. Similarly, high SR is observed in the \emph{SSL}, with values as high as [98, 94], [100, 99], and [97, 96]. The SR observed in the \emph{Alzheimer} is likewise high but relatively lower than in other datasets. For instance, for the scenario of \emph{Hybrid-EE-ECR}, our approach attains an SR of [92, 100], [100, 100], [96, 95]). Particularly, and only for the \emph{Group-ECR} scenario, a relatively low SR is observed with [61.6, 61.1] in \emph{Whole} and [64, 62.5] in C1, which indicates difficulties in generating CFs. Despite these limitations in \emph{Group-ECR} scenario, results suggest that the effectiveness of our approach consistently generates valid CFs that guarantee recourse.

\paragraph*{Success Rate Across Clusters}
We now turn to comparing the SR achieved across \emph{Whole} and the diverse clusters, in each of the scenarios. A noteworthy, thought expected, observation is that the SR values across clusters are consistently higher than those obtained across the \emph{Whole}. For instance, in the \emph{Alzheimer}, in \emph{Group-EE}, SR is [76, 79] across \emph{Whole} but is as high as [88, 91], [94, 100] and [89, 84] in C1, C2 and C3, respectively. Similarly, in \emph{Hybrid-EE-ECR} scenario, SR is [85, 86] for \emph{Whole}, while is [92, 100], [100, 100], and [96, 95] for C1, C2 and C3, respectively. In \emph{Adult}, the results show that the SR in \emph{Whole} is generally either equal to or up to 1–2\% lower than that of the clusters. Except in \emph{Hybrid-EE-ECR} scenario, the SR achieved for \emph{Whole} is [99, 100], while for the 3 clusters is [98, 99]. A clearer understanding of these results will emerge in the subsequent analysis of CF similarity and the overall quality of the generated CFs.

\emph{Takeaway 1: Our approach demonstrates that recourse can be reliably achieved across diverse datasets while sustaining high SR across different demographic groups. This indicates that integrating fairness objectives does not compromise the validity of the generated CFs. Furthermore, our proposed CF generation strategy benefits from operating on more homogeneous subgroups represented by clusters, where the underlying structure and decision boundaries are easier to capture, as opposed to working on the entire dataset, where heterogeneity increases the complexity of generating valid CFs.}


\paragraph*{Proportion Difference}
We now turn to the PD values obtained across groups G1 and G2, reported in Tab. \ref{tab:sr_groups_with_pd} (shown in parentheses) for all scenarios. Recall that PD serves as a fairness metric by quantifying the disparity in SR between the two groups and theoretically ranges from 0 (no disparity) to 100 (complete dispartiy). 

Overall, the results show that PD is relatively low, ranging between 0 and, in worst case, 11.1, which indicates the effectiveness of our approach in ensuring fairness across the two protected groups. More specifically, in \emph{SSL}, our approach keeps PD values within a narrow range of 0–6 across all scenarios and clusters, demonstrating that the CF generation process introduces minimal group disparity. Similarly, in the \emph{Adult}, PD remains between 0–1, while in \emph{Alzheimer} it ranges from 0–11.1. Results show that the reward function in our proposed approach consistently captures PD, ensuring it is kept relatively low across datasets, fairness scenarios, and clusters. It is also noteworthy to highlight that our proposed approach shows comparable performance across all scenarios. 

\emph{Takeaway 2: Our approach not only achieves high validity in terms of SR but also upholds fairness by minimizing disparities between groups. This is reflected in consistently low PD values, which indicate that validity differences across groups are effectively reduced. Consequently, our method guarantees actionable recourse for both individuals and groups, without introducing systematic bias.}

\paragraph*{Impact of Fairness Constraints on SR and PD}
We now turn to comparing the diverse fairness scenarios in each of the datasets in terms of SR and PD. The key question we seek to address is: \emph{To what extent does the introduction of additional fairness constraints (i.e., group and hybrid) influence SR and PD?}. In \emph{Alzheimer}, our approach attains an SR ranging from 75 to 100, and a PD ranging between 0 and 8, across all scenarios, except for \emph{Group-ECR}. Specifically, in \emph{Individual-EE} and \emph{Group-EE}, SR is substantially high (76–100) and PD is consistently low (0–6). In \emph{Hybrid-EE-ECR} we see a more favorable balance with a relatively high SR (ranging between 85 and 100) and a PD between 0 and 8. In contrast, \emph{Group-ECR} shows the lowest SR (61.6 in \emph{Whole} and 62–79 across clusters) and the highest PD (11.1 for C2 and 9.4 for C3). A similar pattern is observed in \emph{Adult} and \emph{SSL}, where \emph{Group-ECR} consistently emerges as the least favorable setting, yielding lower SR and higher PD compared to the other fairness scenarios, yet still within a comparable range.


\emph{Takeaway 3: Group-only fairness constraints (Group-ECR) are associated with reductions in SR and increases in PD across datasets. By contrast, the Hybrid-EE-ECR approach combines individual and group fairness objectives, yielding comparatively high SR while keeping PD at lower levels. These results suggest that multi-level fairness integration offers a more balanced trade-off between performance and equity than group-only constraints.}

\subsection{Fairness in terms of CF Similarity}
So far, we have discussed the SR and PD metrics. The results indicate that the SR values obtained across the different scenarios are highly comparable, while the PD remains relatively low between the two groups in each case. We now turn our attention to analyzing the similarity (measured using Gower distance) between the generated CFs and their corresponding original instances, measured using the Gower distance. This analysis provides a deeper understanding of fairness, as comparable SR values across groups may still conceal disparities, as one group’s CFs might be less similar to the original instances, thereby implying more challenging or less feasible recourse options. This analysis allows us to address the question: \emph{to what extent do CFs preserve fairness not only in terms of their SR but also in the quality and feasibility of the recourse they provide across different groups?}

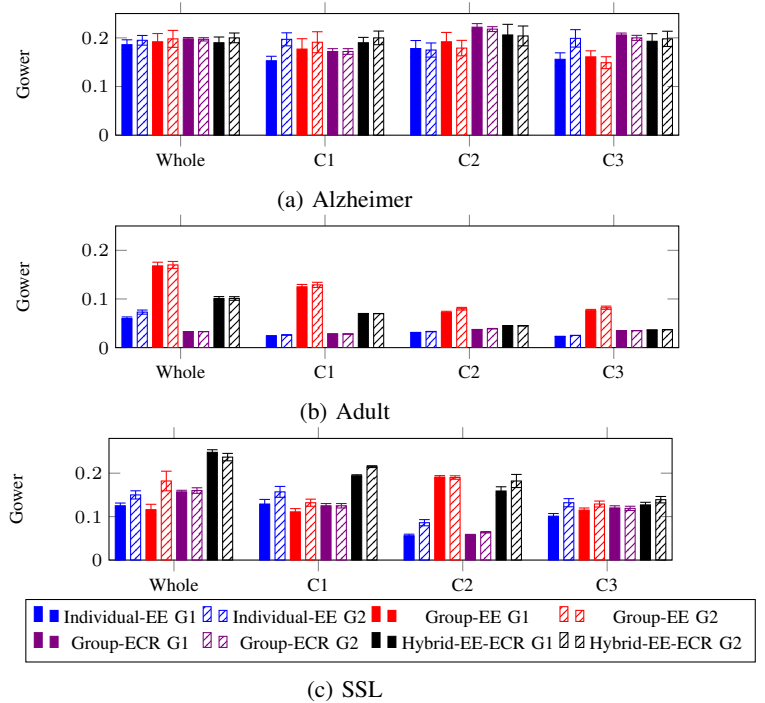
\begin{figure}[t]
\centering
\scriptsize

\begin{subfigure}[t]{0.5\textwidth}
\centering
\begin{tikzpicture}
\begin{axis}[
    ybar,
    bar width=3.8pt,
    enlarge x limits=0.15,
    width=\linewidth,
    height=3.2cm,
    ylabel={Gower},
    xtick=data,
    xticklabels={Whole, C1, C2, C3},
    ymin=0, ymax=0.25,
    tick label style={font=\scriptsize},
    ylabel style={font=\scriptsize},
    label style={font=\scriptsize},
]
\addplot+[blue, fill=blue, error bars/.cd, y dir=both, y explicit relative] coordinates {
    (0,0.186) +- (0,0.054) (1,0.153) +- (0,0.060) (2,0.178) +- (0,0.093) (3,0.156) +- (0,0.085)};
\addplot+[blue, fill=blue, pattern=north east lines, pattern color=blue, error bars/.cd, y dir=both, y explicit relative] coordinates {
    (0,0.195) +- (0,0.051) (1,0.197) +- (0,0.068) (2,0.175) +- (0,0.083) (3,0.199) +- (0,0.090)};
\addplot+[red, fill=red, error bars/.cd, y dir=both, y explicit relative] coordinates {
    (0,0.192) +- (0,0.088) (1,0.177) +- (0,0.120) (2,0.192) +- (0,0.100) (3,0.161) +- (0,0.077)};
\addplot+[red, fill=red, pattern=north east lines, pattern color=red, error bars/.cd, y dir=both, y explicit relative] coordinates {
    (0,0.198) +- (0,0.088) (1,0.191) +- (0,0.113) (2,0.179) +- (0,0.089) (3,0.149) +- (0,0.082)};
\addplot+[violet, fill=violet, error bars/.cd, y dir=both, y explicit relative] coordinates {
    (0,0.198) +- (0,0.014) (1,0.172) +- (0,0.034) (2,0.222) +- (0,0.033) (3,0.206) +- (0,0.019)};
\addplot+[violet, fill=violet, pattern=north east lines, pattern color=violet, error bars/.cd, y dir=both, y explicit relative] coordinates {
    (0,0.197) +- (0,0.017) (1,0.172) +- (0,0.034) (2,0.218) +- (0,0.023) (3,0.200) +- (0,0.026)};
\addplot+[black, fill=black, error bars/.cd, y dir=both, y explicit relative] coordinates {
    (0,0.190) +- (0,0.062) (1,0.190) +- (0,0.058) (2,0.206) +- (0,0.106) (3,0.193) +- (0,0.081)};
\addplot+[black, fill=black, pattern=north east lines, pattern color=black, error bars/.cd, y dir=both, y explicit relative] coordinates {
    (0,0.200) +- (0,0.050) (1,0.200) +- (0,0.070) (2,0.204) +- (0,0.100) (3,0.198) +- (0,0.079)};
\end{axis}
\end{tikzpicture}
\caption{Alzheimer}
\end{subfigure}

\begin{subfigure}[t]{0.5\textwidth}
\centering
\begin{tikzpicture}
\begin{axis}[
    ybar,
    bar width=3.8pt,
    enlarge x limits=0.15,
    width=\linewidth,
    height=3.2cm,
    ylabel={Gower},
    xtick=data,
    xticklabels={Whole, C1, C2, C3},
    ymin=0, ymax=0.25,
    tick label style={font=\scriptsize},
    ylabel style={font=\scriptsize},
    label style={font=\scriptsize},
    legend style={draw=none}
]
\addplot+[blue, fill=blue, error bars/.cd, y dir=both, y explicit relative] coordinates {
    (0,0.060) +- (0,0.052) (1,0.024) +- (0,0.004) (2,0.031) +- (0,0.010) (3,0.023) +- (0,0.005)};
\addplot+[blue, fill=blue, pattern=north east lines, pattern color=blue, error bars/.cd, y dir=both, y explicit relative] coordinates {
    (0,0.073) +- (0,0.058) (1,0.026) +- (0,0.004) (2,0.033) +- (0,0.009) (3,0.025) +- (0,0.004)};
\addplot+[red, fill=red, error bars/.cd, y dir=both, y explicit relative] coordinates {
    (0,0.168) +- (0,0.045) (1,0.125) +- (0,0.039) (2,0.072) +- (0,0.037) (3,0.076) +- (0,0.034)};
\addplot+[red, fill=red, pattern=north east lines, pattern color=red, error bars/.cd, y dir=both, y explicit relative] coordinates {
    (0,0.170) +- (0,0.042) (1,0.129) +- (0,0.041) (2,0.080) +- (0,0.034) (3,0.082) +- (0,0.040)};
\addplot+[violet, fill=violet, error bars/.cd, y dir=both, y explicit relative] coordinates {
    (0,0.033) +- (0,0.001) (1,0.028) +- (0,0.000) (2,0.037) +- (0,0.010) (3,0.035) +- (0,0.003)};
\addplot+[violet, fill=violet, pattern=north east lines, pattern color=violet, error bars/.cd, y dir=both, y explicit relative] coordinates {
    (0,0.033) +- (0,0.001) (1,0.028) +- (0,0.011) (2,0.039) +- (0,0.011) (3,0.035) +- (0,0.003)};
\addplot+[black, fill=black, error bars/.cd, y dir=both, y explicit relative] coordinates {
    (0,0.101) +- (0,0.039) (1,0.070) +- (0,0.003) (2,0.045) +- (0,0.000) (3,0.036) +- (0,0.005)};
\addplot+[black, fill=black, pattern=north east lines, pattern color=black, error bars/.cd, y dir=both, y explicit relative] coordinates {
    (0,0.101) +- (0,0.037) (1,0.070) +- (0,0.002) (2,0.045) +- (0,0.000) (3,0.037) +- (0,0.004)};
\end{axis}
\end{tikzpicture}
\caption{Adult}
\end{subfigure}

\begin{subfigure}[t]{0.5\textwidth}
\centering
\begin{tikzpicture}
\begin{axis}[
    ybar,
    bar width=3.8pt,
    enlarge x limits=0.15,
    width=\linewidth,
    height=3.2cm,
    xtick=data,
    xticklabels={Whole, C1, C2, C3},
    ylabel={Gower},
    ymin=0, ymax=0.28,
    tick label style={font=\scriptsize},
    ylabel style={font=\scriptsize},
    label style={font=\scriptsize},
    legend style={at={(0.5,-0.32)}, anchor=north, legend columns=4, font=\scriptsize}
]
\addplot+[blue, fill=blue, error bars/.cd, y dir=both, y explicit relative] coordinates {
    (0,0.125) +- (0,0.050) (1,0.129) +- (0,0.082) (2,0.056) +- (0,0.064) (3,0.101) +- (0,0.061)};\addlegendentry{Individual-EE G1}
\addplot+[blue, fill=blue, pattern=north east lines, pattern color=blue, error bars/.cd, y dir=both, y explicit relative] coordinates {
    (0,0.150) +- (0,0.063) (1,0.157) +- (0,0.080) (2,0.086) +- (0,0.083) (3,0.132) +- (0,0.071)};\addlegendentry{Individual-EE G2}
\addplot+[red, fill=red, error bars/.cd, y dir=both, y explicit relative] coordinates {
    (0,0.116) +- (0,0.107) (1,0.111) +- (0,0.067) (2,0.191) +- (0,0.018) (3,0.115) +- (0,0.042)};\addlegendentry{Group-EE G1}
\addplot+[red, fill=red, pattern=north east lines, pattern color=red, error bars/.cd, y dir=both, y explicit relative] coordinates {
    (0,0.182) +- (0,0.124) (1,0.132) +- (0,0.062) (2,0.190) +- (0,0.021) (3,0.129) +- (0,0.053)};\addlegendentry{Group-EE G2}
\addplot+[violet, fill=violet, error bars/.cd, y dir=both, y explicit relative] coordinates {
    (0,0.157) +- (0,0.023) (1,0.125) +- (0,0.041) (2,0.058) +- (0,0.020) (3,0.120) +- (0,0.041)};\addlegendentry{Group-ECR G1}
\addplot+[violet, fill=violet, pattern=north east lines, pattern color=violet, error bars/.cd, y dir=both, y explicit relative] coordinates {
    (0,0.160) +- (0,0.040) (1,0.125) +- (0,0.041) (2,0.064) +- (0,0.024) (3,0.119) +- (0,0.041)};\addlegendentry{Group-ECR G2}
\addplot+[black, fill=black, error bars/.cd, y dir=both, y explicit relative] coordinates {
    (0,0.248) +- (0,0.024) (1,0.194) +- (0,0.012) (2,0.159) +- (0,0.061) (3,0.127) +- (0,0.047)};\addlegendentry{Hybrid-EE-ECR G1}
\addplot+[black, fill=black, pattern=north east lines, pattern color=black, error bars/.cd, y dir=both, y explicit relative] coordinates {
    (0,0.237) +- (0,0.036) (1,0.215) +- (0,0.010) (2,0.182) +- (0,0.083) (3,0.139) +- (0,0.053)};\addlegendentry{Hybrid-EE-ECR G2}
\end{axis}
\end{tikzpicture}
\caption{SSL}
\end{subfigure}
\caption{\footnotesize{Gower distance between generated CFs and original instances across the three datasets. Each subplot corresponds to a dataset and shows the similarity for each CF explainer trained on the whole dataset or clusters (C1--C3), separately for protected groups G1 and G2.}}
\label{fig:similarity_all_datasets}
\vspace{-0.5cm}
\end{figure}

Fig. \ref{fig:similarity_all_datasets} shows the average Gower distance between the CFs generated using our approach and the original instances in the three dataset, across the diverse scenarios, clusters, and across protected groups. We first compare the similarity across the diverse scenarios and then we focus on the group granularity.

\paragraph*{CF Similarity Across Scenarios}
Overall, the values of Gower distance range approximately between 0 and 0.2 (with 0 being similar and 1 dissimilar). This indicates that the generated CFs are similar to their respective original instances in the feature space. For instance, in the \emph{Adult}, in \emph{Individual-EE}, Gower distances as low as 0.025 are observed. Similarly, in \emph{SSL}, Gower distances are mostly below 0.20. Additionally, we note that the variance across scenarios is generally low. For example, in the \emph{Alzheimer} and \emph{Group-ECR} scenario, the observed Gower distance is consistently around 0.198 with very low standard deviation (e.g., $\pm 0.014$ and $\pm 0.017$ for G1 and G2, respectively). Importantly, we observe that \emph{Hybrid-EE-ECR} maintains a low Gower distance and generates CFs similar to original instances as much as those generated in \emph{Individual-EE} and \emph{Group-ECR} when considered separately. 

\paragraph*{Impact of Clustering on CF Similarity}
Comparing CFs Gower distance values between \emph{Whole} and the three clusters, we observe that the CFs in \emph{Whole} generally exhibit equal or slightly higher distances than those in the clusters. This trend is shared across all scenarios and datasets. For instance, in \emph{Adult}, for \emph{Group-EE}, the average Gower distance for \emph{Whole} is 0.19, while that of the diverse clusters ranges from 0.10 to 0.17. Also in \emph{SSL}, for \emph{Hybrid-EE-ECR}, the average Gower distance observed for \emph{Whole} is around 0.22, higher than those of C1, C2 and C3, which range from 0.15 to 0.20. These findings suggest that clustering generally yields CFs that are more tightly grouped in the feature space, leading to greater internal consistency compared to those derived from the \emph{Whole}. 

\paragraph*{Group Fairness in terms of CF Similarity}
We now examine the extent to which our proposed approach promotes fairness in the feasibility of recourse by comparing the Gower distance achieved for G1 and G2. Results show that across all scenarios, the difference between the Gower distances of G1 and G2 is relatively small and, in most cases, even negligible. This pattern is consistent also across \emph{Whole} and the diverse clusters. Specifically, in \emph{Alzheimer}, the differences between CF Gower of groups are negligible, around 0.01, with the largest difference appearing in cluster C3 in the \emph{Individual-EE} scenario, where it is around 0.04, which is still considered relatively low. A similar trend is observed for \emph{Adult} and \emph{SSL}, for instance, \emph{Group-ECR} scenario of \emph{Adult}, G1 and G2 exhibit nearly identical distances of 0.033 and \emph{SSL} for \emph{Hybrid-EE-ECR} scenario, the Gower for G1 and G2 in cluster C3 are 0.127 and 0.139, respectively, showing a minimal difference.

These results lead us to the following two key takeaways:
\emph{Takeaway 4: The generated CFs at the cluster level produce more precise and similar outcomes compared to generating them for the entire dataset. This improvement stems from the ability to customize CFs based on subgroup-specific patterns, resulting in more targeted solutions with reduced distance from original instances and fairness across groups, as reflected by a comparable CF similarity across groups.}

\emph{Takeaway 5: The \emph{Hybrid-EE-ECR} approach maintains CF similarity while balancing individual- and group-level objectives, reinforcing its role as a principled formulation of hybrid fairness. The generated CFs remain close to the original data points, as reflected by consistently low Gower distances, making them both effective and actionable solutions. This closeness is preserved across groups, ensuring fairness due to minimal differences in CF similarity.}

\subsection{Number of Actions}
\begin{figure*}[t]
\centering
\scriptsize

\begin{subfigure}[b]{0.33\textwidth}
\centering
\begin{tikzpicture}
\begin{axis}[
    ybar,
    bar width=5pt,
    enlarge x limits=0.15,
    width=\linewidth,
    height=3.3cm,
    ylabel={Nb Actions},
    xtick=data,
    xticklabels={Whole, C1, C2, C3},
    ymin=0, ymax=5,
    tick label style={font=\scriptsize},
    ylabel style={font=\scriptsize},
    label style={font=\scriptsize},
    legend style={
        at={(1.8,1.3)},          
        anchor=south,          
        legend columns=2,
        font=\small,      
        draw=none,
        /tikz/every even column/.append style={column sep=1em},
    },
]
\addplot+[violet, fill=violet] coordinates {(0,1) (1,2) (2,3) (3,2)};
\addlegendentry{Group-ECR}
\addplot+[black, fill=black] coordinates {(0,3) (1,3) (2,2) (3,3)};
\addlegendentry{Hybrid-EE-ECR}
\end{axis}
\end{tikzpicture}
\caption{Alzheimer}
\end{subfigure}
\begin{subfigure}[b]{0.33\textwidth}
\centering
\begin{tikzpicture}
\begin{axis}[
    ybar,
    bar width=5pt,
    enlarge x limits=0.15,
    width=\linewidth,
    height=3.3cm,
    xtick=data,
    xticklabels={Whole, C1, C2, C3},
    ymin=0, ymax=5,
    tick label style={font=\scriptsize},
    ylabel style={font=\scriptsize},
    label style={font=\scriptsize},
    legend style={draw=none},
]
\addplot+[violet, fill=violet] coordinates {(0,2) (1,2) (2,3) (3,2)};
\addplot+[black, fill=black] coordinates {(0,4) (1,3) (2,2) (3,2)};
\end{axis}
\end{tikzpicture}
\caption{Adult}
\end{subfigure}
\begin{subfigure}[b]{0.33\textwidth}
\centering
\begin{tikzpicture}
\begin{axis}[
    ybar,
    bar width=5pt,
    enlarge x limits=0.15,
    width=\linewidth,
    height=3.3cm,
    xtick=data,
    xticklabels={Whole, C1, C2, C3},
    ymin=0, ymax=5,
    tick label style={font=\scriptsize},
    ylabel style={font=\scriptsize},
    label style={font=\scriptsize},
    legend style={draw=none},
]
\addplot+[violet, fill=violet] coordinates {(0,4) (1,1) (2,3) (3,1)};
\addplot+[black, fill=black] coordinates {(0,1) (1,2) (2,2) (3,1)};
\end{axis}
\end{tikzpicture}
\caption{SSL}
\end{subfigure}
\caption{\footnotesize{Number of actions required to reach a recourse for protected groups across the three datasets. Since G1 and G2 yield identical values, only one bar per method is displayed. Results are shown for the Group-ECR and Hybrid-EE-ECR methods across the whole dataset and clusters (C1-C3).}}
\label{fig:nb_actions_all_datasets_ECR}
\end{figure*}
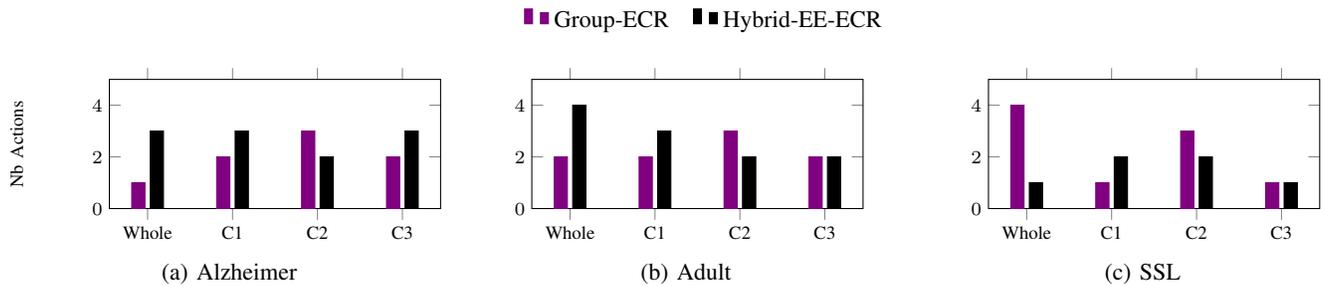

We now focus our analysis on the number of generated actions for \emph{Group-ECR} and \emph{Hybrid-EE-ECR}, as these scenarios are directly optimizing for the diversity of CFs. This analysis allows to determine if the ECR constraint is consistently met by ensuring equal number of choices between the protected groups and, consequently, if our approach is capable of ensuring diversity in the CFs by providing many recourse options. Fig. \ref{fig:nb_actions_all_datasets_ECR} shows the number of actions achieved for \emph{Whole} and the diverse clusters in the \emph{Group-ECR} and \emph{Hybrid-EE-ECR}, across the three datasets. Note that we show only one bar per case (scenario-cluster), as the number of actions for G1 and G2 resulted the same in each of the cases. First, we highlight that the fact that the number of actions resulted the same for both groups across all cases indicates that our approach was capable of identifying an equal choice of recourse for individuals to choose from, which proves the effectiveness of our approach in providing and ensuring fair and equitable recourse choices, regardless of which group the individual belongs to.

\paragraph*{Impact of Clustering on the Number of Actions}
We now move to discussing how the clustering impacts the number of actions (available recourse options). This is relevant as it reveals how clustering, as expected, can increase the total availability of recources across the datasets. A clear and consistent trend emerges: the clustered approach significantly increases the total number of unique actions compared to when considering the dataset as a \emph{Whole}, across all datasets. This is because the clustering identifies distinct local recourse options, allowing for the generation of tailored CF options for different subgroups that the \emph{Whole} scenario would often miss. For example, in \emph{Alzheimer}, our approach identified 3 actions for \emph{Whole} while for C1, C2 and C3, it identified 3, 2,and 3, respectively, resulting in a total of 8 distinct recourse options. 

\emph{Takeaway 6: Clustering enables the exploitation of local data patterns to identify flexible and varied recourse options. The approach also maintains fairness by guaranteeing that protected groups receive equivalent numbers of actionable CFs, balancing recourse diversity with fair access.}

\paragraph*{Comparing Group-ECR to Hybrid-EE-ECR}
We now analyze the impact of introducing fairness constraints through \emph{Hybrid-EE-ECR} on the number of recourse options, compared to that of \emph{Group-ECR}. This will allow us to answer \emph{Does optimizing for \emph{Hybrid-EE-ECR} impact the number of available recourse options, with respect to only optimizing for \emph{Group-ECR}?} The results reveal an interesting and non-uniform trend. Contrary to the expectation that the \emph{Hybrid-EE-ECR} approach would generate fewer actions, our findings show a variable relationship across different clusters and datasets. This can be attributed to the nature of the optimization problem: The \emph{Hybrid-EE-ECR} scenario imposes additional fairness constraints such as those of individual-level fairness, hence, it effectively narrows the search space for valid CFs. In some cases, \emph{Hybrid-EE-ECR} leads to a simpler, more precise solution that satisfies both individual and group fairness criteria simultaneously. In contrast, the \emph{Group-ECR}, with its broader objective, may find a more complex CF that works for an entire group but is not necessarily the minimal for any single individual. These findings are consistent with the results presented in Tab. \ref{tab:sr_groups_with_pd} and Fig. \ref{fig:similarity_all_datasets}, where the \emph{Group-ECR} exhibited a lower SR and a higher Gower distance compared to the \emph{Hybrid-EE-ECR}.

\emph{Takeaway 7: \emph{Hybrid-EE-ECR} introduces no additional complexity in generating CFs in term of the number of actions, when compared to \emph{Group-ECR}. However, \emph{Group-ECR} tends to yield more complex CFs that deviate further from the original instances and are associated with lower SR and higher PD.}


\subsection{Comparative Evaluation Against Baseline Approaches}

\begin{table*}[htbp]
\renewcommand{\arraystretch}{1.15} 
\setlength{\tabcolsep}{6pt} 
\centering
\caption{CF quality metrics across datasets and scenarios. Values are shown for [G1, G2] with absolute difference in parentheses. For reference, training set plausibility is 0.822 (Alzheimer), 0.382 (Adult), and 0.15 (SSL).}
\label{tab:cf_quality_full_clean_diff}
\begin{scriptsize}
\begin{tabular}{llcccc}
\toprule
\textbf{Dataset} & \textbf{Scenario} & \textbf{Validity} & \textbf{Plausibility} & \textbf{Similarity} & \textbf{Minimality} \\
\midrule
\multirow{7}{*}{\textbf{Alzheimer}} 
  & Individual-EE     & [91.9\%, 93.1\%] (1.2) & [0.848, 0.800] (0.048) & [0.192, 0.220] (0.028) & [2.608, 2.687] (0.079) \\
  & Group-EE          & [93.0\%, 91.7\%] (1.3) & [0.852, 0.799] (0.053) & [0.204, 0.207] (0.003) & [2.725, 2.788] (0.063) \\
  & Group-ECR         & [83.7\%, 81.9\%] (1.8) & [0.854, 0.803] (0.051) & [0.223, 0.225] (0.002) & [2.778, 2.661] (0.117) \\
  & \textbf{Hybrid-EE-ECR} & \textbf{[96.5\%, 98.6\%]} (2.1) & \textbf{[0.850, 0.799]} (0.051) & \textbf{[0.203, 0.210]} (0.007) & \textbf{[3.000, 3.000]} (0.000) \\
  & NiCE              & [100\%, 100\%] (0.0) & [0.876, 0.825] (0.051) & [0.405, 0.419] (0.014) & [1.547, 1.472] (0.075) \\
  & Prototype         & [91.9\%, 91.7\%] (0.2) & [0.867, 0.826] (0.041) & [0.201, 0.179] (0.022) & [16.785, 16.106] (0.679) \\
  & DiCE              & [100\%, 100\%] (0.0) & [0.870, 0.822] (0.048) & [0.365, 0.365] (0.000) & [1.267, 1.278] (0.011) \\
\midrule
\multirow{7}{*}{\textbf{Adult}} 
  & Individual-EE     & [97.4\%, 98.5\%] (1.1) & [0.468, 0.392] (0.076) & [0.071, 0.073] (0.002) & [1.085, 1.148] (0.063) \\
  & Group-EE          & [99.3\%, 99.6\%] (0.3) & [0.480, 0.405] (0.075) & [0.115, 0.121] (0.006) & [2.745, 2.703] (0.042) \\
  & Group-ECR         & [97.4\%, 98.5\%] (1.1) & [0.470, 0.392] (0.078) & [0.053, 0.056] (0.003) & [1.918, 1.868] (0.050) \\
  & \textbf{Hybrid-EE-ECR} & \textbf{[97.9\%, 99.1\%]} (1.2) & \textbf{[0.474, 0.397]} (0.077) & \textbf{[0.081, 0.081]} (0.000) & \textbf{[2.000, 2.000]} (0.000) \\
  & NiCE              & [100\%, 100\%] (0.0) & [0.549, 0.628] (0.079) & [0.211, 0.197] (0.014) & [1.213, 1.275] (0.062) \\
  & Prototype         & [53.8\%, 37.5\%] (16.3) & [0.483, 0.408] (0.075) & [0.406, 0.427] (0.021) & [6.116, 6.421] (0.305) \\
  & DiCE              & [100\%, 100\%] (0.0) & [0.529, 0.467] (0.062) & [0.249, 0.245] (0.004) & [1.577, 1.559] (0.018) \\
\midrule
\multirow{7}{*}{\textbf{SSL}} 
  & Individual-EE     & [92.1\%, 91.8\%] (0.3) & [0.050, 0.150] (0.100) & [0.093, 0.136] (0.043) & [1.133, 1.408] (0.275) \\
  & Group-EE          & [98.5\%, 96.9\%] (1.6) & [0.190, 0.220] (0.030) & [0.191, 0.192] (0.001) & [1.992, 1.921] (0.071) \\
  & Group-ECR         & [91.4\%, 88.6\%] (2.8) & [0.060, 0.100] (0.040) & [0.116, 0.151] (0.035) & [1.103, 1.157] (0.054) \\
  & \textbf{Hybrid-EE-ECR} & \textbf{[92.1\%, 91.8\%]} (0.3) & \textbf{[0.260, 0.320]} (0.060) & \textbf{[0.280, 0.264]} (0.016) & \textbf{[1.135, 1.414]} (0.279) \\
  & NiCE              & [100\%, 100\%] (0.0) & [0.540, 2.140] (1.600) & [0.051, 0.101] (0.050) & [1.142, 1.312] (0.170) \\
  & Prototype         & [71.8\%, 40.1\%] (31.7) & [0.420, 1.550] (1.130) & [0.034, 0.041] (0.007) & [1.006, 1.046] (0.040) \\
  & DiCE              & [91.5\%, 95.9\%] (4.4) & [1.410, 2.240] (0.830) & [0.211, 0.231] (0.020) & [1.560, 1.693] (0.133) \\
\bottomrule
\end{tabular}
\label{tab:baselines}
\end{scriptsize}
\end{table*}

We now compare the quality of CFs generated by our diverse scenarios of our approach to that of baseline approaches, namely, NICE, DICE, and prototype-guided, in terms of validity, plausibility, similarity, and minimality, as reported in Tab. \ref{tab:baselines}. This analysis is relevant as it clarifies the implications of enforcing fairness constraints compared to traditional CFs.

\paragraph*{Validity}
Across all datasets and scenarios, our approach consistently achieves high CF validity, typically ranging from 91.4\% to 98.6\%, with the exception of \emph{Alzheimer} under \emph{Group-ECR}. This performance is comparable to the validity of CFs generated with DiCE and NiCE, and more stable than prototype-based CFs, which display substantial drops in certain datasets. For instance, our approach under \emph{Hybrid-EE-ECR} scenario attains validity rates of 96.5–98.6\% on \emph{Alzheimer}, 97.9–99.1\% on \emph{Adult}, and around 92\% on \emph{SSL}, for both groups. DiCE reaches full validity (100\%) on \emph{Alzheimer} and \emph{Adult}, but performs slightly lower on \emph{SSL}. By contrast, prototype-based CFs exhibit highly inconsistent behavior and achieve 91.7\% on \emph{Alzheimer}, while their validity is low on \emph{Adult} (as low as 37.5\%) and \emph{SSL} (40.1\%) for certain groups. 

We now compare the results achieved across the diverse fairness scenarios. For \emph{Individual-EE} and \emph{Group-EE}, validity remains very high generally in the upper 90s across \emph{Alzheimer} and \emph{Adult}, and above 91\% for \emph{SSL}. In contrast, the \emph{Group-ECR} scenario yields slightly lower validity in some cases (e.g., 81.9\% on \emph{Alzheimer}, 88.6\% on \emph{SSL}), reflecting the stricter constraints imposed by enforcing ECR. Importantly, our approach under the \emph{Hybrid-EE-ECR} scenario preserves validity at consistently high levels, ranging from 96.5–98.6\% on \emph{Alzheimer}, 97.9–99.1\% on \emph{Adult}, and around 92\% on \emph{SSL}. These results demonstrate that our approach maintains consistently high CF validity across datasets and fairness scenarios, with \emph{Hybrid-EE-ECR} achieving both robustness and stability relative to baselines and group constraints.

\paragraph*{Plausibility} 
In terms of plausibility, measured as reconstruction error (Sec. \ref{ProblemFormulationAndMehtodology}), where lower values indicate more plausible CFs, our approach across fairness scenarios achieves plausibility scores that are consistently close to the training reconstruction error, indicating that the generated CFs remain realistic and aligned with the data distribution. For example, in the \emph{Alzheimer}, plausibility ranges between 0.799 and 0.854 compared to the training error of 0.822, with similar patterns observed on \emph{Adult} and \emph{SSL}. Compared to the baselines, a significant difference emerges: both NiCE and DiCE achieve very high validity at the expense of plausibility, yielding reconstruction errors that are noticeably higher than those of our approach ($>$ 0.86). Prototype-based CFs show competitive plausibility in some cases, but this comes at the cost of substantially lower validity, making them less reliable overall.

\paragraph*{Similarity and Minimality}
Similarity and minimality assess how closely generated CFs resemble the original instances. NiCE and DiCE produce CFs with relatively high Gower distances compared to the fairness scenarios, for example, 0.365–0.419 on \emph{Alzheimer} and 0.211–0.249 on \emph{Adult}, indicating less similarity to the original inputs. Prototype CFs perform even worse, with distances as high as 0.406–0.427 on \emph{Adult}, they alter up to 16 features on \emph{Alzheimer} and $>6$ on \emph{Adult}. In contrast, our approach consistently produces CFs that are both closer to the original instances and require only a small number of modifications across all scenarios. For example, in \emph{Individual-EE} in \emph{Adult}, the approach generates CFs with Gower distances of 0.071–0.073 while modifying just over a single feature. In \emph{Group-EE}, our proposed approach results in a slightly higher number of feature changes (2.7 features on \emph{Adult}) but still maintains low Gower values (0.115–0.121). In \emph{Hybrid-EE-ECR}, the approach provides a balanced trade-off, producing CFs with moderate edits (e.g., two features and a similarity of 0.081 on \emph{Adult}, and 3 features with similarity values of 0.203–0.210 on \emph{Alzheimer}). Overall, across datasets, our approach preserves proximity to the original data while avoiding excessive feature edits.

\emph{Takeaway 8: Our approach demonstrates that fairness constraints do not necessarily impose significant drawbacks on CF quality. While baseline approaches often exhibit trade-offs between fairness and performance metrics, such as achieving high validity at the expense of similarity. The integration of fairness objectives preserves validity, plausibility, and similarity at levels comparable to or exceeding existing baselines. In addition, hybrid fairness balances individual and group fairness while preserving the quality of CFs and maintaining fairness at both levels.}

\subsection{Proof of Convergence of RL}
\begin{figure}[h]
\centering
    \begin{subfigure}{0.33\textwidth}
        \centering
        \begin{tikzpicture}
            \begin{axis}[
                xlabel=Time-step, 
                ylabel=Episode reward mean, 
                width=0.8\linewidth,
                height=0.35\linewidth,
                enlargelimits=false,
                scale only axis=true
            ]
                \addplot [mark=none] table [col sep=comma, x=Step, y=Value] {Files/EF-ECR_Rew.csv};
            \end{axis}
        \end{tikzpicture}
        \caption{Episode Reward Mean}
    \end{subfigure}%
    
    \begin{subfigure}{0.33\textwidth}
        \centering
        \begin{tikzpicture}
            \begin{axis}[
                xlabel=Time-step, 
                ylabel=Entropy coefficient, 
                width=0.8\linewidth, 
                height=0.35\linewidth,
                enlargelimits=false,
                scale only axis=true
            ]
                \addplot [mark=none] table [col sep=comma, x=Step, y=Value] {Files/EF-ECR_Entropy.csv};
            \end{axis}
        \end{tikzpicture}
        \caption{Entropy Coefficient}
    \end{subfigure}
    \caption{Visualization of RL performance metrics for a sample experiment on the Alzheimer dataset.}
    \label{ExampleConvergenceAlzheimer}
\end{figure}
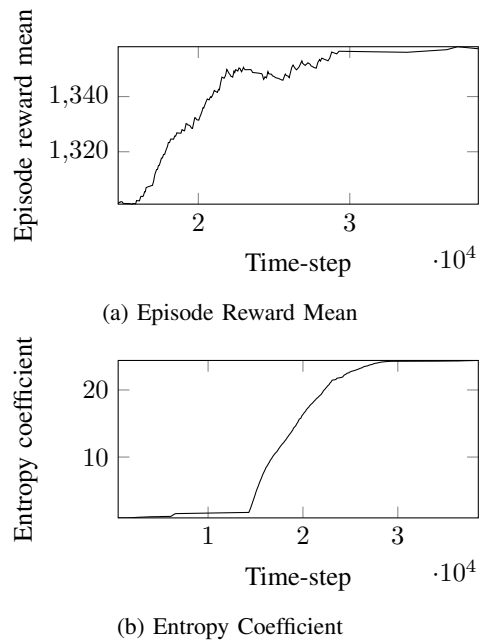

Fig. \ref{ExampleConvergenceAlzheimer} illustrates the convergence plots of the RL-based approach with Hybrid-EE-ECR during training, showing the evolution of two key SAC metrics: mean reward and the entropy coefficient. The mean reward curve shows that the RL agent's performance improves over time, indicating effective learning that generates CFs. The entropy coefficient plot reflects the degree of exploration during training. The rising mean reward indicates that the agent is learning to maximize cumulative rewards. The upward trend in the entropy coefficient confirms that the agent balances reward maximization with exploration, a desirable property for robust policy learning and the search of better CF solutions. A key limitation of RL is its high computational complexity, which often results in long training times and substantial resource requirements.

\section{Conclusion}
In this work, we address the problem of generating fair counterfactual explanations (CFs) with the aim of ensuring that individuals who share comparable attributes or belong to different protected groups (e.g., demographic groups), are provided with comparable explanations. We define fairness in CFs at individual, group, and hybrid levels and propose a model-agnostic reinforcement learning based approach. Our approach is designed to optimize CF fairness by employing customized reward functions that enforce equal effectiveness and equal choice of recourse, thereby balancing fairness with the quality of the generated CFs. Experiments conducted on three datasets demonstrate the effectiveness of our approach in generating CFs while adhering to fairness constraints and preserving their quality. Additionally, we demonstrate the trade-offs of applying hybrid fairness constraints and show that it does not inherently lead to compromises in CF quality.

\bibliographystyle{IEEEtran}
\bibliography{paper}
\end{document}